\documentclass[10.5pt]{article}

\usepackage{wrapfig}

\usepackage{mathpazo}
\usepackage[utf8]{inputenc}
\usepackage[T1]{fontenc}
\usepackage{verbatim}
\usepackage{calc}
\usepackage{mathrsfs}
\usepackage{mathtools}
\usepackage{url}
\usepackage{algorithm2e}
\usepackage{amsmath}
\usepackage{amssymb}
\usepackage{graphicx}
\usepackage{hyperref}
\usepackage{xcolor}
\usepackage{comment}
\usepackage{soul}
\usepackage{booktabs}
\usepackage{tikz}
\usetikzlibrary{matrix,arrows.meta,positioning}
\usepackage[most]{tcolorbox}

\definecolor{forestgreen}{rgb}{0.13, 0.55, 0.13}

\usepackage{tabularx}

\makeatletter

\newcommand{\lyxmathsym}[1]{\ifmmode\begingroup\def\b@ld{bold}
  \text{\ifx\math@version\b@ld\bfseries\fi#1}\endgroup\else#1\fi}

\usepackage{graphics}\usepackage{epsfig}

\newcommand\hc{ \rowcolor{teal!10}}
\newcommand\hd{ \rowcolor{teal!18}}

\usepackage{soul}

\usepackage{graphicx}
\usepackage{booktabs} 

\usepackage{here}

\usepackage{amsmath,amssymb,amsfonts,amsbsy,amsfonts,latexsym}
\usepackage{multirow}
\usepackage{makecell}
\usepackage[labelfont=bf,belowskip=0pt,aboveskip=5pt,tableposition=top]{caption}
\usepackage{xcolor}
\usepackage{colortbl}

\definecolor{colorA}{RGB}{189,201,225}
\definecolor{colorB}{RGB}{103,169,207}
\definecolor{colorC}{RGB}{ 28,144,153}
\definecolor{colorD}{RGB}{  1,108, 89}

\newcolumntype{R}{>{\columncolor{gray!40}}r}
\newcolumntype{L}{>{\columncolor{gray!40}}l}
\newcolumntype{C}{>{\columncolor{gray!40}}c}

\usepackage{tabularx,colortbl,xcolor}
\usepackage{multirow}
\usepackage[normalem]{ulem}
\useunder{\uline}{\ul}{}

\usepackage{enumitem}

\usepackage{xparse}

\DeclareGraphicsExtensions{.pdf,.png}

\SetKwInput{KwInput}{Input}

\usepackage{longtable}
\usepackage{pgfplots}
\usepackage{outlines}

\usepackage{caption}
\usepackage{subcaption}
\usepackage{graphbox} 

\tcbset{
    sharp corners,
    colback = white,
    before skip = 0.2cm,    
    after skip = 0.5cm      
}                           

\definecolor{main}{HTML}{4472C4}    
\definecolor{sub}{HTML}{EBF4FF}     

\newtcolorbox{boxA}{
    enhanced, breakable,
    boxrule = 0pt,
    colback = sub,
    borderline west = {2pt}{0pt}{main}, 
    borderline east = {2pt}{0pt}{main}, 
}

\usepackage{times}
\usepackage{textcomp}

\newcommand{\OURS}{{\textsc{XQuant}}\xspace}
\newcommand{\OURSCL}{{\textsc{XQuant-CL}}\xspace}

\title{
\OURS: Breaking the Memory Wall for \\
LLM Inference with KV Cache Rematerialization
}

\author{
Aditya Tomar\thanks{Equal Contribution}\hspace{0.5em}$^{1}$\enspace\enspace Coleman Hooper\footnotemark[1]\hspace{0.5em}$^{1}$\enspace\enspace Minjae Lee$^{2}$\enspace\enspace  Haocheng Xi$^{1}$\\
Rishabh Tiwari$^{1}$ \enspace\enspace Wonjun Kang$^{2}$ \enspace\enspace Luca Manolache$^{1}$\\
Michael W. Mahoney$^{1,3,4}$\enspace\enspace Kurt Keutzer$^{1}$\enspace\enspace Amir Gholami$^{1,3}$\vspace{3mm}
\\
{$^{1}$~UC Berkeley\qquad $^{2}$~FuriosaAI\qquad $^{3}$~ICSI\qquad $^{4}$~LBNL\vspace{3mm}}\\
}
\date{}  
\makeatother

\begin{document}

\maketitle

\begin{abstract}

Although LLM inference has emerged as a critical workload for many downstream applications, efficiently inferring LLMs is challenging due to the substantial memory footprint and bandwidth requirements.
In parallel, compute capabilities have steadily outpaced both memory capacity and bandwidth over the last few decades, a trend that remains evident in modern GPU hardware and exacerbates the challenge of LLM inference.
As such, new algorithms are emerging that trade increased computation for reduced memory operations.
To that end, we present \OURS, which takes advantage of this trend, enabling an order-of-magnitude reduction in memory consumption through low-bit quantization with substantial accuracy benefits relative to state-of-the-art KV cache quantization methods.
We accomplish this by quantizing and caching the layer input activations \textit{X}, instead of using standard KV caching, and then rematerializing the Keys and Values on-the-fly during inference. 
This results in an immediate 2$\times$ memory savings compared to KV caching.
By applying \OURS, we achieve up to $\sim 7.7\times$ memory savings with $<0.1$ perplexity degradation compared to the FP16 baseline.
Furthermore, our approach leverages the fact that \textit{X} values are similar across layers. 
Building on this observation, we introduce \OURSCL, which exploits the cross-layer similarity in the \textit{X} embeddings for extreme compression. 
Across different models, \OURSCL attains up to 10$\times$ memory savings relative to the FP16 baseline with only 0.01 perplexity degradation, and 12.5$\times$ memory savings with only $0.1$ perplexity degradation. 
Notably, despite using standard uniform quantization, \OURSCL is able to surpass intricate KV cache quantization methods that employ non-uniform quantization with outlier-aware strategies.
Given the aforementioned trends in compute versus memory scaling for future generations of hardware platforms, \OURS adopts a forward-looking perspective to accelerate LLM inference: \OURS seeks to exploit the rapidly increasing compute capabilities to eliminate the memory bottleneck, while surpassing state-of-the-art KV cache quantization methods and achieving near-FP16 accuracy across a wide range of models.

\end{abstract}

\section{Introduction}

\begin{figure}[t]
    \centering
    \includegraphics[width=1\linewidth]
    {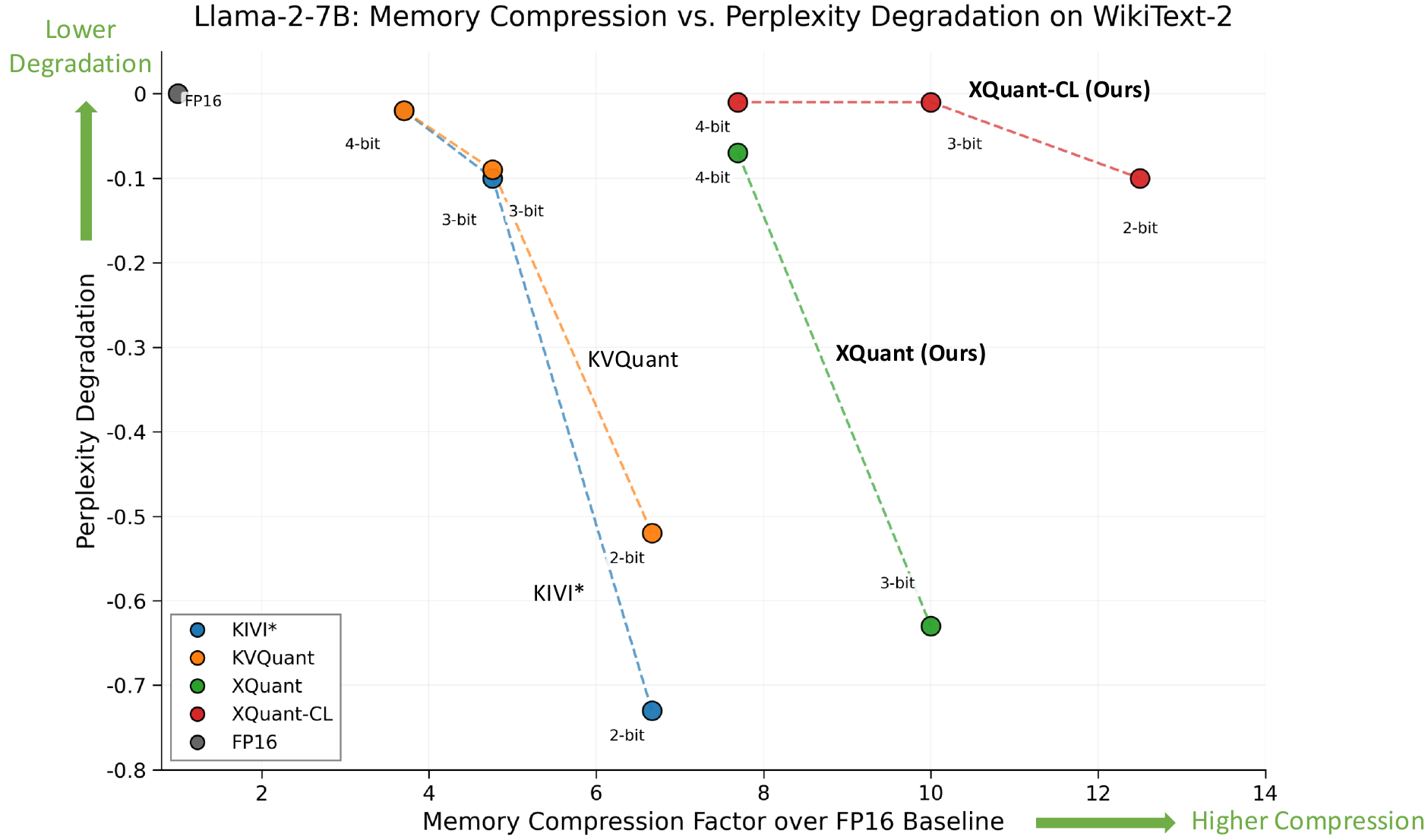}
    \caption{Perplexity degradation (lower is better) versus memory compression factor (higher is better) evaluated using Llama-2-7B on WikiText-2 for state-of-the-art KV cache quantization methods and for our \OURS, across \{4,3,2\}-bit widths. The top right edge of the plot represents the optimal configuration that attains the most memory compression and the least perplexity degradation. Memory compression factor and perplexity degradation are with respect to the FP16 baseline. As shown in Table \ref{tab:xquant-cl-ppl}, \OURSCL achieves only 0.01 perplexity degradation while getting $10\times$ memory savings with 3-bit quantization, and 0.1 perplexity degradation while getting $12.5\times$ memory compression with 2-bit quantization.}
    \label{fig:memory-compression-ppl}
\end{figure}

\begin{figure}[t]
\centering
\includegraphics[width=1\linewidth]{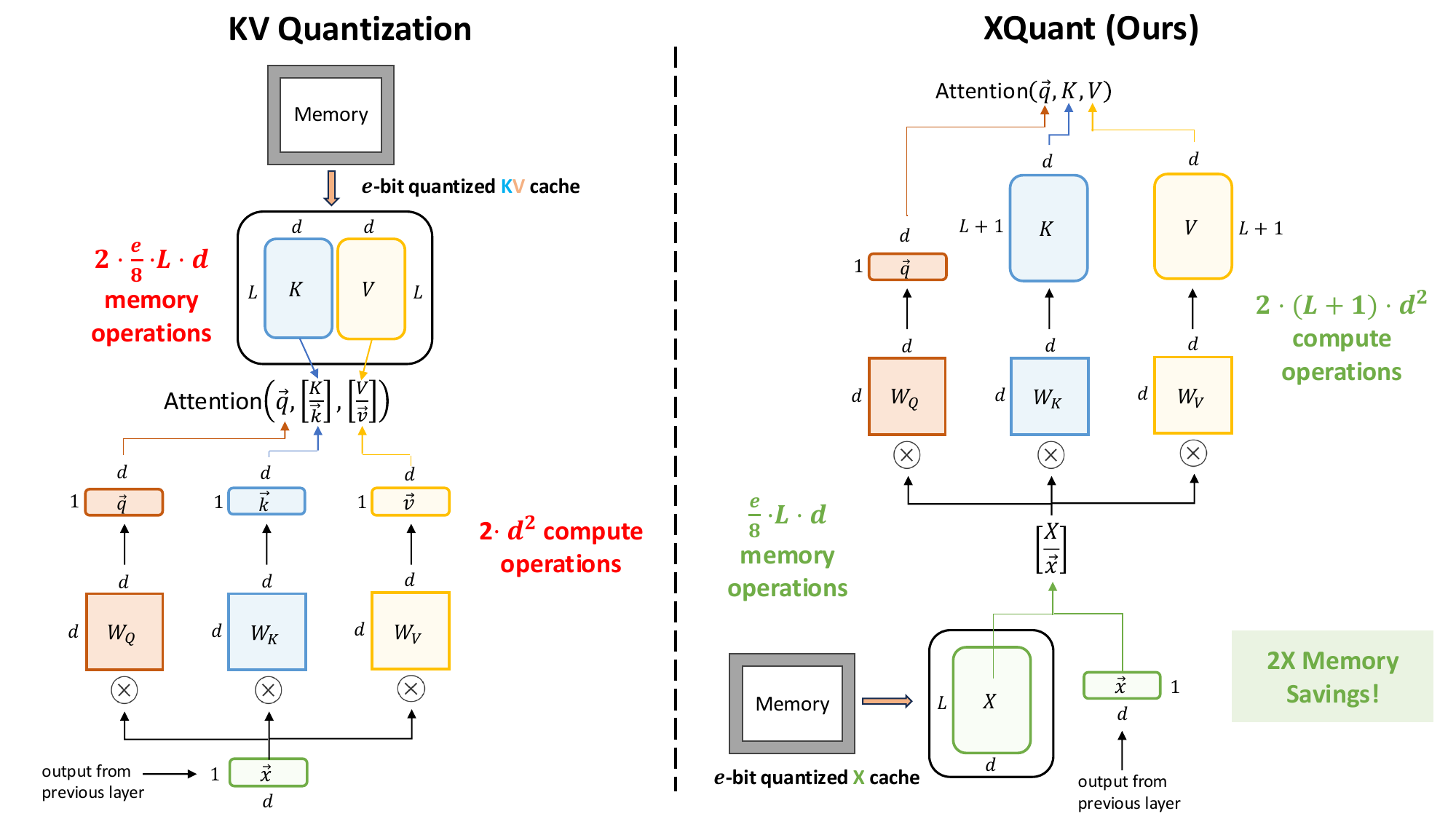}
 \caption{
 A visualization of how \OURS reduces the memory footprint by caching the input embedding (\textit{X}) instead of the KV cache.
 We use the cached input to rematerialize the Keys and Values in order to compute attention.
 This increases the amount of computation required when computing attention. However, since LLM inference is typically memory bandwidth-bound, we can accelerate inference by reducing memory operations, even at the expense of additional compute operations.
 }
  \label{fig:thumbnail}
\end{figure}

Large Language Models (LLMs) have seen widespread adoption as a standard paradigm across a range of Natural Language Processing (NLP) applications \cite{chowdhery2023palm,brown2020language,team2023gemini,touvron2023llama}.
While LLMs achieve remarkable performance on these tasks, they have substantial inference costs due to their large parameter count as well as the number of memory operations required when running generation.
Prior work has demonstrated how LLM inference tends to be \textit{Memory Bandwidth-Bound}, rather than Compute-Bound \cite{kim2024squeezellm, tiwari2025quantspec, sadhukhan2025magicdecbreakinglatencythroughputtradeoff}, and therefore reducing the memory footprint of LLMs is critical to enable downstream applications. 
For short context lengths and small batch sizes, the model weights are typically the memory bottleneck. 
However, for long context lengths and large batch sizes, the main memory bottleneck for LLM inference is the Key-Value (KV) cache, which is the embedded representation of the entire sequence used in the self-attention mechanism and which grows linearly with respect to the sequence length \cite{tiwari2025quantspec, sadhukhan2025magicdecbreakinglatencythroughputtradeoff}. 
During inference, generating each new token requires repeatedly loading and storing the entire KV cache, which becomes prohibitively expensive and leads to substantial slowdown.
This motivates efforts to reduce KV cache memory operations to speed up the inference process.
One promising solution to compress the KV cache is through KV cache quantization \cite{hooper2024kvquant, liu2024kivi}.
Quantizing the KV cache reduces the memory footprint and number of memory operations required during decoding by using fewer bits to represent the Keys and Values. 
However, while existing methods retain accuracy even when quantizing the KV cache to low precision (e.g., 4-bit quantization), further reducing the bit-width of KV activations often degrades model performance.

In this work, we present \OURS, a method which quantizes the input activations \textit{X} of each layer, rather than the KV cache, to reduce the required memory consumption.
Our method is visualized in Figure \ref{fig:thumbnail}.
Quantizing \textit{X} provides a 2$\times$ memory savings compared with quantizing the KV cache, since we only need to store one tensor per layer instead of separate Keys and Values. Interestingly, we also find that \textit{X} is more amenable to extremely low-bit quantization than the KV cache. Moreover, although rematerializing KV cache from $X$ requires additional computation during decoding, we can afford this cost because LLM inference is progressively becoming more memory-bandwidth bound. 
This phenomenon will be increasingly prevalent with future hardware platforms, as the rate of improvement in compute capabilities continue to outpace increases in memory bandwidth and capacity \cite{gholami2024ai}.
In this work, we make the following contributions (which are summarized in Figure \ref{fig:memory-compression-ppl}):

\begin{itemize}
    \item To reduce the memory consumption for LLM inference, \OURS quantizes the input \textit{X} activations, providing a 2$\times$ memory savings relative to quantizing the KV cache directly, and then rematerializes the KV activations on-the-fly during inference (see Section \ref{sec:algo-xquant}). 
    We show that for the same memory footprint ($\sim 7.7\times$ savings compared to FP16), \OURS attains up to $\sim0.9$ less perplexity degradation compared to KV cache quantization and $< 0.1$ perplexity degradation compared to the FP16 baseline (see Section \ref{sec:xquant-ppl}).
    \item For ultra-low precision quantization with accuracy comparable to the FP16 baseline, we present \OURSCL, a method that exploits the cross-layer similarity in the \textit{X} embeddings. Our approach compresses the differences in \textit{X} between successive layers, which have much smaller range as a result of the residual stream in the Transformer\cite{vaswani2017attention} architecture (see Section \ref{sec:cross-layer}). Using standard asymmetric uniform quantization, we observe only $0.01$ perplexity degradation with 3-bit quantization while attaining $10 \times$ memory savings compared to the FP16 baseline, and only $0.1$ perplexity degradation with 2-bit quantization and $12.5\times$ memory savings relative to FP16 (see Section~\ref{sec:xquant-cl-ppl}). 
    Remarkably, \OURSCL outperforms state-of-the-art KV cache quantization methods like KVQuant \cite{hooper2024kvquant} that use complex techniques such as non-uniform quantization and outlier-aware dense-and-sparse strategies. 
    On the Wikitext-2 and C4 datasets, we reduce perplexity degradation by $\sim 0.4$ compared to KVQuant while using $1.9\times$ less memory (see Section~\ref{sec:xquant-cl-ppl}).
    \item We extend \OURS and \OURSCL to support new models that use Grouped Query Attention (GQA) \cite{ainslie2023gqa}.
    For GQA models, we decompose the Key and Value projection weight matrices offline using the singular value decomposition (SVD), and we then down-project the input activations into a smaller latent space, reducing memory consumption. Interestingly, we observe that the \textit{X} distribution in this latent space is suitable for extremely low precision quantization (see Section \ref{sec:algo-xquant-svd}). 
    For 2-bit quantization, \OURS achieves less than 2.2 perplexity degradation relative to KV cache quantization (see Section \ref{sec:xquant-ppl}). 
    \OURSCL further improves performance by resulting in only $\sim0.1$ perplexity degradation compared to the FP16 baseline while saving $6.7\times$ memory (see Section \ref{sec:xquant-cl-ppl}).
    \item Finally, we provide a system-level analysis of computational overheads and memory savings from rematerialization. We show that \OURS is able to significantly  reduce memory consumption while preserving accuracy, which will ultimately result in speedup even at the cost of additional computation, as compute continues to dominate memory capacity and bandwidth (see Section \ref{sec:modeling}).
\end{itemize}

\section{Related Work}

\subsection{Memory Wall for LLM Inference}
\label{sec:memory-wall}
When assessing the performance of kernels on a target hardware platform, their runtime is determined by either 1) the amount of compute operations that need to be performed or 2) the amount of memory operations that need to be performed. 
Which of these two factors is the bottleneck depends on the characteristics of the target hardware as well as those of the kernel.
\textit{Arithmetic Intensity}, which is defined as the ratio of the number of compute operations that need to be performed per byte transferred to or from memory, is a typical metric to evaluate a kernel and is usually expressed in units of floating point operations (FLOPs) per byte:
\begin{equation}
    \mathrm{Arithmetic\ Intensity} = \frac{\mathrm{Compute\ Operations\ Performed}}{\mathrm{Bytes\ Transfered}} ,
\end{equation}
We can characterize the target hardware platform in terms of the ratio of the peak compute performance of the device versus the amount of memory bandwidth that it provides. 
The ratio of peak compute to memory bandwidth, typically also expressed as FLOPs per byte similar to arithmetic intensity, is referred to here as the \textit{ridge point} for that hardware platform (we use this term as it aligns with the point on a Roofline plot \cite{williams2009roofline} which delineates compute-bound versus memory-bandwidth bound kernels for a target hardware platform):
\begin{equation}
    \mathrm{Ridge\ Point} = \frac{\mathrm{Peak\ Compute\ Throughput}}{\mathrm{Peak\ Memory\ Bandwidth}} .
\end{equation}

When assessing whether a provided kernel is memory bandwidth-bound or compute-bound, we need only compare the arithmetic intensity of the kernel with the ridge point of our target hardware platform.
If the arithmetic intensity is larger, then the kernel is compute-bound when run on the target hardware; if the arithmetic intensity is smaller, then the kernel will be memory-bandwidth bound~\cite{tiwari2025quantspec,williams2009roofline,yuan2024llm}.

When considering LLM inference, a key challenge is that it has low arithmetic intensity for most batch sizes and context length regimes \cite{kim2023full,kim2024squeezellm,tiwari2025quantspec}. 
This is because generating each token only requires performing low intensity matrix-vector multiplications, whereas loading these matrices requires far more memory operations. 
Thus, LLM inference performs very few floating point operations per byte loaded from memory.
Additionally, when we assess scaling trends with LLMs and hardware compute and memory, there is a growing discrepancy between the memory capabilities of existing devices and an LLM's demands.
This phenomenon has been termed the Memory Wall problem
\cite{gholami2024ai}. 
There are two key components to this problem:
\begin{enumerate}
    \item The scaling trends in terms of memory requirements for modern LLMs have dramatically outpaced both the increases in memory capacity and bandwidth on hardware.
    \item The scaling in peak computational performance is orders of magnitude greater than the corresponding increases in memory capacity and bandwidth on hardware.
\end{enumerate}
Taking these factors together, it is critical to reduce the number of memory operations required for LLM inference. 
If we can additionally reduce the memory requirements by increasing the amount of computation, this will be beneficial, due to the growing gap 
between peak computational performance and memory capabilities of hardware platforms.

\begin{tcolorbox}[colback=blue!10!white, colframe=blue!60!black, title=Motivation]
While increasing computation in exchange for reduced memory usage may introduce latency on today’s hardware, this tradeoff is expected to become increasingly favorable. Our work leverages this trend, aiming to reduce memory bottlenecks by utilizing additional compute, ultimately enabling faster LLM inference on future hardware generations.
\end{tcolorbox}

\subsection{Activation Rematerialization}

Prior work has explored rematerializing activations, where activations are recomputed on-the-fly from a smaller checkpointed state, as a partial solution to the memory wall problem \cite{jain2020checkmate}. 
Checkmate \cite{jain2020checkmate} allowed training large DNNs in memory-constrained GPUs by retaining a subset of intermediate activations as checkpoints, with other activations being discarded and then rematerialized from these checkpointed states.
Checkmate solves for the optimal recomputation setup for the target hardware platform with provided memory constraints.
Recent work has also explored rematerializing the KV cache from the input embedding \textit{X} in the context of serving systems which offload some of the KV entries to the CPU.
HCache \cite{gao2025fast} offloads \textit{X} to CPU memory, and performs restoration by moving \textit{X} from CPU to GPU and then recomputing to recover the Keys and Values. \cite{lee2025efficient} also performs rematerialization to recover the Keys and Values from \textit{X}, and determines the optimal amount of rematerialization to perform, given a combination of compute and memory constraints. 
While these works focus on system-level optimizations to determine the amount of rematerialization that can be performed, our work focuses on compressing \textit{X} rather than KV cache activations to attain improved savings relative to existing KV cache compression methods, and then exploits cross-layer similarity in \textit{X} to attain greater memory savings.

There has also been prior work on trying to avoid storing Keys and Values separately for the KV cache.
El-Attention \cite{yan2021attention} merged the Key and Value projection matrices into other matrices in the model, thereby allowing for computing attention directly using the input embedding.
However, this method is not compatible with the rotary positional embedding (RoPE) encodings or with models that use Grouped Query Attention.
Slim-Attention \cite{graef2025slim} similarly aimed to only cache Keys and to multiply them by the inverse Key projection matrix to recover the values (and to merge this inverse matrix into other matrices in the model offline).
This approach also requires applying RoPE on-the-fly during inference and is not compatible with models that use grouped-query attention. 
Moreover, such inverses are not guaranteed to be numerically stable.

\subsection{KV Cache Quantization}

KV cache quantization has emerged as a promising method for reducing the memory requirements for the KV cache by using fewer bits to represent each floating point element in each KV cache entry.
Previous work has quantized the Key distributions per-channel and the Value distributions per-token in order to adapt to the outlier channels in Keys \cite{hooper2024kvquant,liu2024kivi}.
A crucial aspect of Key cache quantization is handling RoPE \cite{su2024roformer}.
Prior methods have either applied pre-RoPE Key quantization \cite{hooper2024kvquant} or quantized the Key cache using polar-form representations \cite{wu2025polarquant,han2025polarquant}, in order to retain accuracy.
Mixed-precision KV cache quantization has been used in order to preserve model accuracy by retaining particularly sensitive tokens in higher precision \cite{yang2024no,he2024zipcache}.
Prior work has also explored retaining initial pivot tokens intact \cite{liu2024intactkv}; this builds on prior work that identified initial tokens as ``attention sink,'' tokens which are disproportionately important for preserving model accuracy \cite{xiao2023efficient}.

\subsection{Low-Rank Decomposition for KV Cache and KV Rematerialization}

There has also been prior work which has applied a low-rank decomposition to the KV cache or the corresponding projection matrices, in order to cache KV entries with a reduced latent dimension before rematerializing the original KV entries.
xKV \cite{chang2025xkv} exploits the fact that the singular vectors for the KV cache entries for successive layers are well aligned to group the KV cache entries across layers and apply SVD to the concatenated KV caches. Loki \cite{singhania2024loki} finds that the keys are low-rank and performs attention in a lower dimension to identify the most important keys, and then only loads those keys for full-dimensional sparse attention. While Loki only reduces memory operations, Eigen Attention \cite{saxena2024eigen} actually compresses the KV cache and only does attention in low-rank space.
There have also been multiple works which have applied low-rank decomposition to the weights in order to project the KV cache to a lower dimension \cite{zhang2024lorc,chang2024palu,yan2025recalkv}.
LoRC \cite{zhang2024lorc} applies SVD to the weight matrices and keeps more singular values at earlier layers to minimize error amplification.
Palu \cite{chang2024palu} performs a low-rank decomposition on the weights offline ahead of inference, and then caches the intermediate KV cache entries.
ReCalKV \cite{yan2025recalkv} reorders heads before applying SVD to groups of heads in order to retain accuracy and reduce rematerialization overhead. 
In contrast with these works that aim to compress the KV cache using a low-rank decomposition, \OURS aims to quantize the \textit{X} embeddings in order to get a 2$\times$ memory savings without applying a low-rank decomposition.
\OURSCL also exploits the cross-layer similarity in the \textit{X} distributions by compressing the differences across layers, which provides substantial memory reduction for the same accuracy relative to compressing the KV cache directly.

\section{Algorithm: Sacrificing Compute to Alleviate Memory Bottlenecks}
\label{sec:algorithm}
In this section, we introduce our algorithms \OURS (Section \ref{sec:algo-xquant}) and \OURSCL (Section \ref{sec:cross-layer}), and we discuss them in the context of Multi-Head Attention (MHA) models \cite{vaswani2017attention}. Then in Section \ref{sec:algo-xquant-svd}, we discuss how we extend our algorithm to support Grouped Query Attention (GQA) models \cite{ainslie2023gqa}. Note that we specifically address GQA because of its widespread popularity and adoption as an optimization technique for KV cache reduction among many model families such as Llama, Mistral, Gemma, Qwen, etc. \cite{dubey2024llama, jiang2023mistral, gemmateam2025gemma3technicalreport, yang2025qwen3technicalreport}. Lastly, in Section \ref{sec:modeling}, we discuss the tradeoffs between compute and memory operations that our methods make.

\subsection{\OURS: Quantizing \textit{X} Instead of KV}
\label{sec:algo-xquant}

The core idea of \OURS is to reduce the memory requirements of KV caching by checkpointing the input activations and regenerating the Keys and Values from these smaller checkpoints when they need to be used to compute attention.
This is shown in Figure \ref{fig:thumbnail}, where we quantize and cache the input embedding \textit{X}, which requires 2$\times$ less memory than using standard KV caching.
The drawback with caching \textit{X} is that it requires rematerializing the KV cache on-the-fly; this requires multiplying the input embedding by projection matrices $W_k$ and $W_v$.
However, as outlined in Sections \ref{sec:memory-wall} and \ref{sec:modeling}, since the arithmetic intensity is typically low for attention in LLM decoding, we can afford to perform additional computation for rematerialization in order to reduce the number of memory operations required. 
Note that whenever we refer to \textit{X}, we mean the input activations after layer normalization has been applied to them \cite{vaswani2017attention, zhang2019rootmeansquarelayer, xiong2020layernormalizationtransformerarchitecture}.

\subsection{\OURSCL: Leveraging Cross-Layer Similarity in \textit{X}}


\label{sec:cross-layer}

\begin{figure}[t]
\centering
\includegraphics[width=1\linewidth]{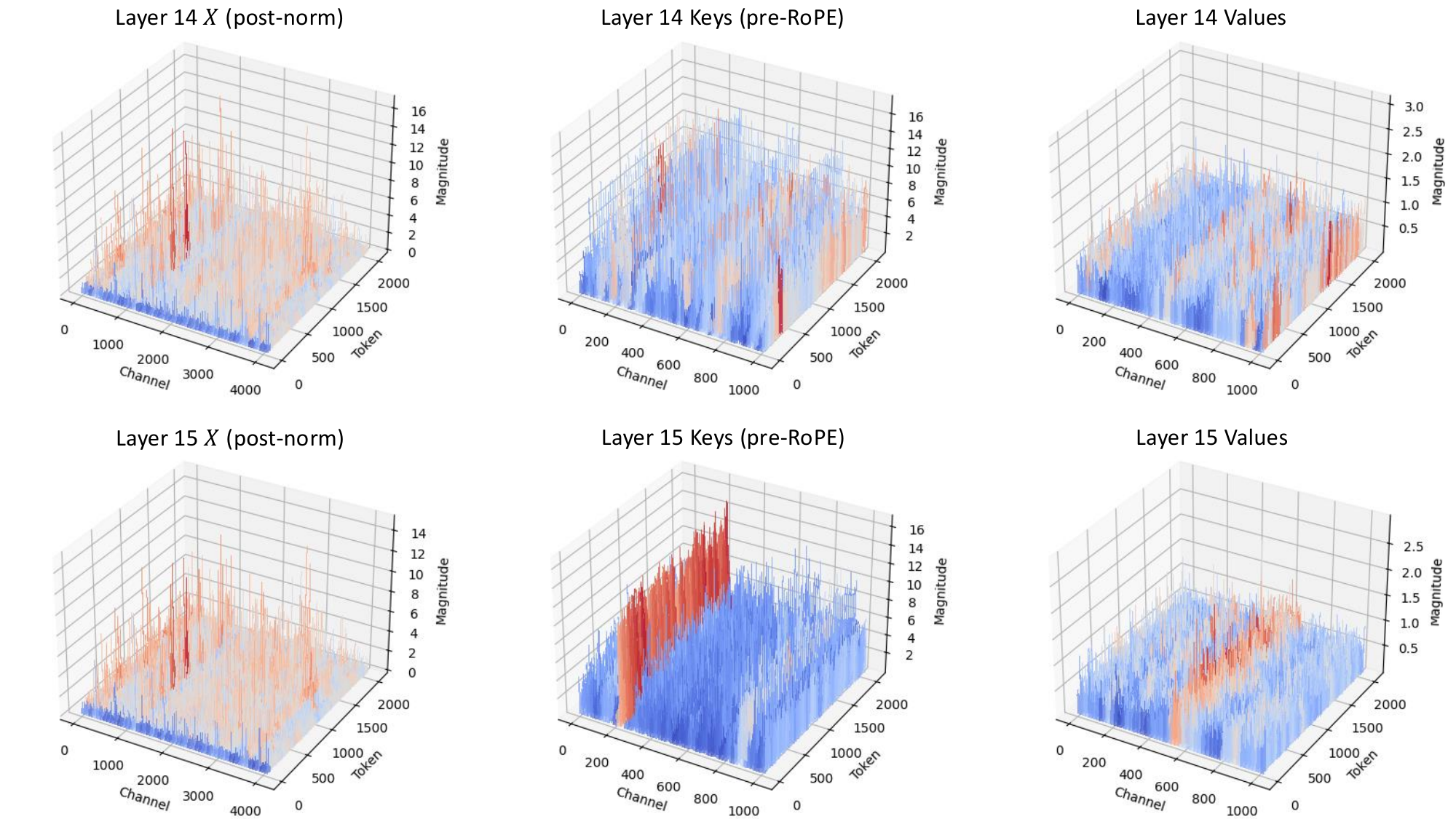}
 \caption{
 Comparison of the post-norm input embeddings \textit{X}, pre-RoPE Keys, and Values for successive layers in the Llama-3.1-8B model.
 The distributions were collected using a test sample with 2K sequence length from Wikitext-2.
 Although the Keys and Values exhibit distinct differences across successive layers, the \textit{X} embeddings bear remarkable similarity. We exploit this similarity using cross-layer compression in \OURSCL.
}
  \label{fig:cross-layer-distn}
\end{figure}

As shown in Figure \ref{fig:cross-layer-distn}, the \textit{X} embeddings across successive layers are remarkably similar (when compared with the similarities of KV cache embeddings across layers).
This observed property can be attributed to the residual stream which flows from each layer's input to the outputs of the attention and multi-layer perceptron blocks within that layer, where these outputs
are added to the residual stream \cite{vaswani2017attention}. 
In this way, each layer's function can be understood as simply refining its input, and since this refinement occurs gradually, it is intuitive
that the inputs of successive layers are not substantially dissimilar \cite{jastrzebski2018residual}.
This represents a promising opportunity for further compression if we can exploit cross-layer similarity.

\begin{figure}[t]
\centering
\includegraphics[width=1\linewidth]{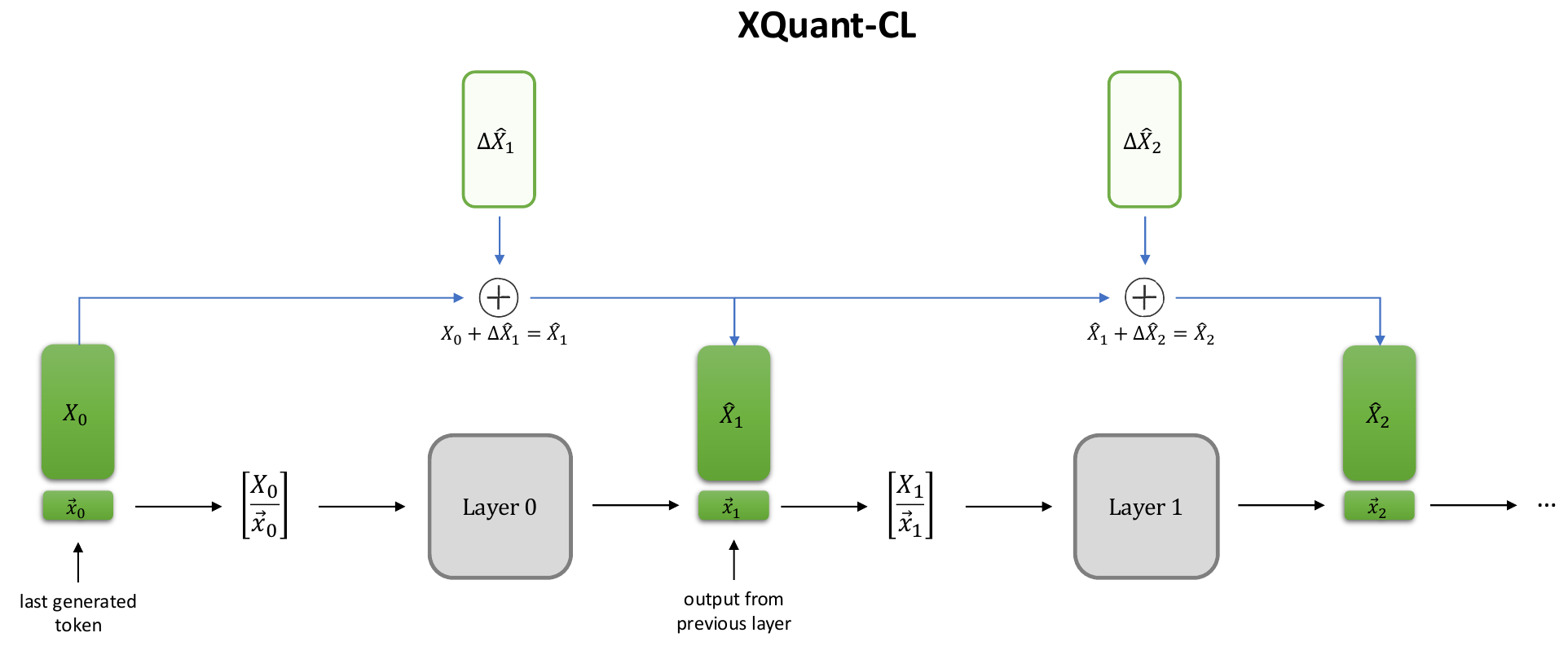}
 \caption{
 Illustration of \OURSCL algorithm during decoding. Besides Layer 0, the input to all other layers is a cross layer approximation, computed using the deltas of all previous layers and the input of Layer 0. The input of Layer 0 is summed with each layer's delta so it can be treated as an accumulator, allowing us to avoid loading all $N-1$ deltas to compute Layer $N$'s $X$. After a layer is done processing, the input embedding to the layer for the last token is subtracted from the output activations of the layer (the same shape as a single token), and this delta is quantized and appended to the $\Delta \hat{X}$ cache. \OURSCL during prefill is visualized in Figure \ref{fig:cross-layer-prefill} in Appendix \ref{sec:appendix-xquant-cl}, which shows how the full $\Delta \hat{X}$ is computed and cached for each layer.
}
  \label{fig:cross-layer-decode}
\end{figure}

To exploit the cross-layer similarity in \textit{X}, we propose \OURSCL, which compresses the \textit{differences} between \textit{X} for successive layers. Our algorithm during prefill is illustrated in Figure \ref{fig:cross-layer-prefill} in Appendix \ref{sec:appendix-xquant-cl}, and our algorithm for decoding is visualized in Figure \ref{fig:cross-layer-decode}.
Instead of directly quantizing \textit{X} for layer $i$, we instead quantize $\Delta X_i = X_i - X_{i-1}$, which has a much smaller range and hence is much easier to quantize.
For this method, we leave the first layer $X_0$ in higher precision, and then compute the differences relative to $X_0$.
However, if we directly compute $X_i - X_0$ for a layer $i$ which is far from layer $0$, $X_i$ would have accumulated enough changes such that it has diverged substantially from $X_0$, meaning that this delta is no longer easy to quantize.
To address this, we quantize and cache $\Delta \hat{X}_i = Q(X_i - \hat{X}_{i-1})$, where $Q$ is the quantization function and $\hat{X}_{i-1}$ is the cross-layer approximation for the previous layer's $X$. 
We then approximate $X_i$ as $\hat{X}_i = X_0 + \sum_{j=1}^i \Delta \hat{X}_j$.
This sequential delta summation allows for the explicit accumulation of the refinements that each layer applies to its input, allowing us to exploit the easy quantization properties of these individual deltas for extreme compression. However with the above formulation, computing $\hat{X}$ for layer $i$ requires loading all $i-1$ previous deltas, which is very expensive. Thus, we maintain an accumulator which sums the deltas of all previous layers (as highlighted in Figure \ref{fig:cross-layer-decode}). This way, computing $\hat{X}_i$ only requires loading the accumulator ($\hat{X}_{i-1} = X_0 + \sum_{j=1}^{i-1} \Delta \hat{X}_j$) and a single delta $\Delta\hat{X}_i$.

\subsection{Support for Grouped-Query Attention Models}
\label{sec:algo-xquant-svd}

\begin{figure}[t]
\centering
\includegraphics[width=1\linewidth]{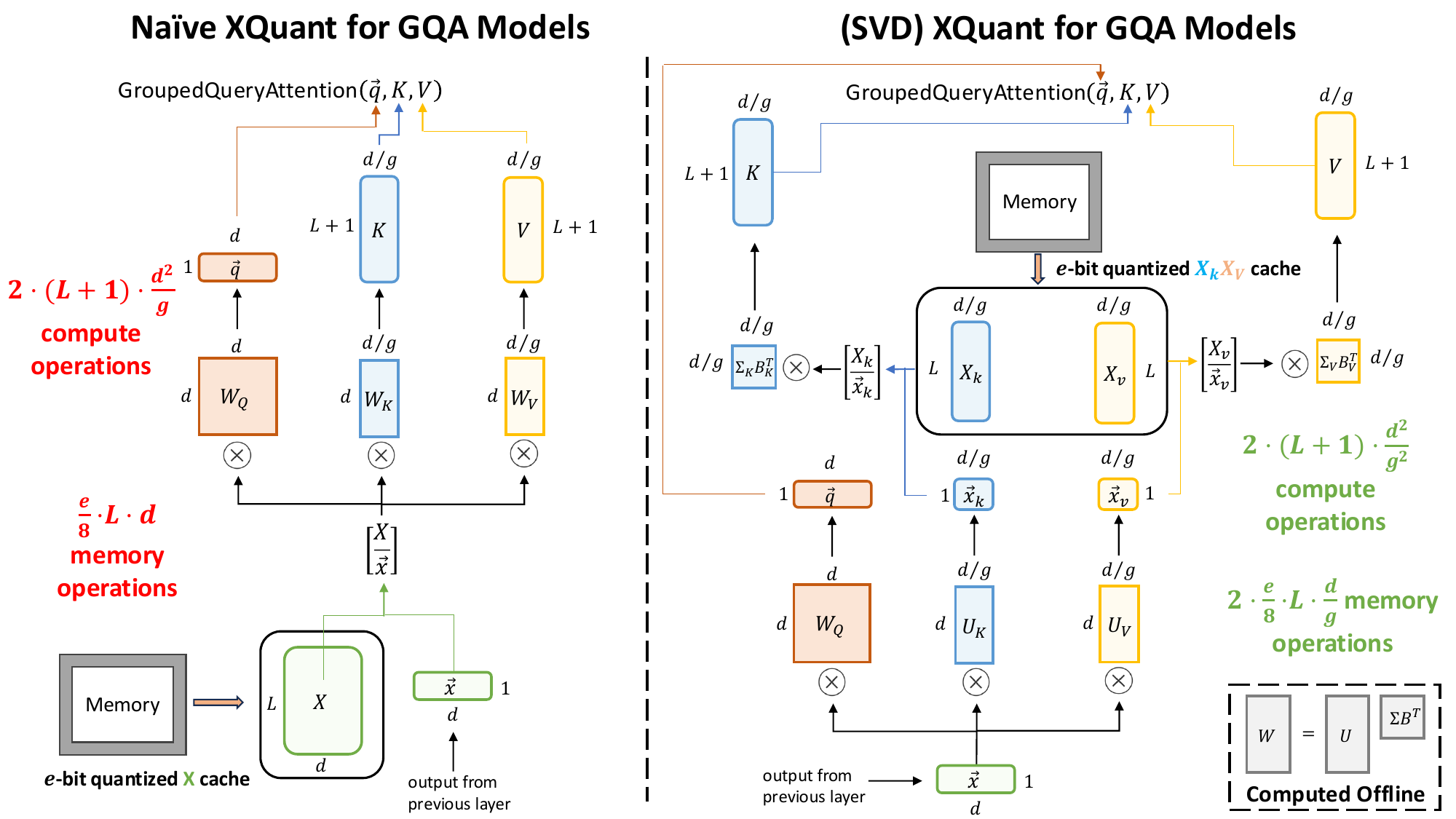}
 \caption{
 A diagram outlining how we apply \OURS for GQA-based models.
 GQA down-projects the input embedding (\textit{X}) to a smaller $d/g$ dimension when computing the Keys and Values. Hence, if we naively quantize the input \textit{X} rather than the KV cache, this will potentially have greater memory consumption.
 To address this, we first apply SVD to the $W_k$ and $W_v$ matrices offline.
 Online during prefill, we down-project $X$ by $U_k$ to get $X_k$ and by $U_v$ to get $X_v$ before applying \OURS, thereby reducing memory consumption. For generation, each new token is also down-projected into this latent space, appended to the $X_k$ and $X_v$ cache, and quantized. The concatenated $[X_k|\vec{x}_k]$ is multiplied by ($\Sigma_kB^T_k$) to recompute the Keys, and the concatenated $[X_k|\vec{x}_v]$ is multiplied by ($\Sigma_vB^T_v$) to recompute the values. Note that the group size $g$ is typically greater than or equal to 4, meaning that naively caching \textit{X} uses greater than or equal to $2\times$ as much memory as simply doing KV caching.
 }
  \label{fig:svd}
\end{figure}

One challenge with extending \OURS to many newer LLMs is that these models typically use Grouped Query Attention (GQA) \cite{ainslie2023gqa}, as opposed to standard Multi-Headed Attention (MHA) \cite{vaswani2017attention} used by the models we've addressed thus far. 
As mentioned previously, we explicitly address GQA models since GQA has seen broad adoption for LLMs as a way of reducing the size of the KV cache \cite{dubey2024llama, jiang2023mistral, gemmateam2025gemma3technicalreport, yang2025qwen3technicalreport}.
GQA reduces the memory consumption of the KV cache by sharing Keys and Values amongst a subset of the attention heads.
The challenge with extending our methodology to GQA models is that they down-project the input activation when computing the Keys and Values. 
This means that while the \textit{X} embeddings have shape $l\times d$ (where $l$ is the sequence length and $d$ is the hidden dimension), K and V are $l\times \frac{d}{g}$ each, where $g$ is the number of query heads that share Keys and Values.
For example, the Llama-3.1-8B model\cite{llama31} has a hidden dimension of 4,096 and uses $g=4$. Therefore, \textit{X} has dimension $l \times 4K$ and the Key and Value activations have dimension $l \times 1K$ each, so the KV cache is of shape $l \times 2K$.
Caching KV would therefore require storing 2 vectors of size 1K each for each token, whereas caching \textit{X} would require storing a single vector of size 4K for each token.
This means that if we naively apply \OURS to GQA models, we will have 2$\times$ memory overhead for the same precision, which negates the benefits of our approach.

\subsubsection{Extending \OURS to support GQA}
\label{sec:xquant-gqa}

To apply \OURS for models which use GQA, we apply a Singular Value Decomposition (SVD) offline to the weight matrices in order to allow the input activations to be stored in a lower-dimensional latent space.
Our algorithm is visualized in Figure \ref{fig:svd}.
We apply SVD to the $W_k$ and $W_v$ matrices to obtain $U_k \Sigma_k B_k^{T}$ and $U_v \Sigma_v B_v^{T}$, respectively. 
During prefill, we down-project the input embeddings as $XU_k$ and $XU_v$ (where $U_k$ and $U_v$ are each $d \times \frac{d}{g}$), which lowers the dimensionality of $X$ by $\frac{g}{2}$ and results in the same memory footprint as the GQA KV cache. 
During generation, each new token's input activation $\vec{x}$ is similarly down-projected by $U_k$ and $U_v$ to get $\vec{x}_k$ and $\vec{x}_v$, respectively. 
These are appended to the cached $XU_k$ and $XU_v$, which are then multiplied by $\Sigma_k B_k^{T}$ and $\Sigma_v B_v^{T}$, respectively, to get the Keys and Values.
Moreover, $\Sigma_k B_k^{T}$ and $\Sigma_v B_v^{T}$ are each fused ($\Sigma$ and $B^T$ are multiplied to merge them into a single matrix)
to serve as the new weight matrices that the latent $U_k$ and $U_v$, respectively, are multiplied with to rematerialize the KV cache. 
This also reduces the recomputation cost during inference, as the fused $\Sigma_k B_k^{T}$ and $\Sigma_v B_v^{T}$ are smaller square matrices of shape $\frac{d}{g} \times \frac{d}{g}$. We elaborate on the associated cost in more detail in Section \ref{sec:modeling}.
Importantly, the SVD and weight fusing are done \textit{offline} and therefore don't add any latency overhead.

Importantly, note that while this approach has the same memory consumption as KV quantization for the same bit width, the down-projected \textit{X} distributions are easier to quantize, giving us higher accuracy for the same bit width (see Section \ref{sec:xquant-ppl}). In fact, the latent $X$ distributions reveal a very interesting structure: we find that the latent $XU_k$ groups all outliers on the first channel, and we observe this for all layers of the model across several different models on different datasets (see Figures \ref{fig:llama-latent-x-distributions}, \ref{fig:mistral-latent-x-distributions}). Note that we quantize and cache $XU_k$ without applying the $\Sigma_k$ matrix of singular values. $U_k$ is a matrix with orthonormal columns, so observing this structure in $XU_k$ where all the outliers lie on the first channel is very interesting. We do not, however, observe any similar interesting structure in the $XU_v$ distribution. This is in keeping with observations made by other works which find that Keys have outlier channels whereas Values do not have a clearly structured axis where outliers lie \cite{liu2024kivi, hooper2024kvquant}. We believe that the outliers present in the first channel of $XU_k$ are simply distributed to other channels once the latent distribution is transformed by $B^T_v$, which gives rise to the outlier channels found in the Keys. We discuss this further in Appendix \ref{sec:appendix-xquant-gqa}. We utilize per-channel quantization for $XU_k$ and per-token quantization for $XU_v$, as we find this to be the best configuration resulting in the least accuracy degradation. This is similar to \cite{liu2024kivi, hooper2024kvquant} which also quantize the Keys per-channel and Values per-token.

\subsubsection{Extending \OURSCL to support GQA}
\label{sec:xquant-cl-gqa}

We also extend our cross-layer method to support GQA models. For GQA models, \OURSCL faces the same issue addressed in Section $\ref{sec:xquant-gqa}$, specifically that GQA models down-project \textit{X} of shape $l \times d$ to the KV cache of shape $l \times 2 \frac{d}{g}$. Since $\Delta X$ has the same shape as $X$, naively caching this delta for GQA models results in more memory overhead than storing the KV cache. To address this, we column-wise concatenate the $W_k$ and $W_v$ projection matrices, resulting in $W_{kv} = [W_k | W_v] \in \mathbb{R}^{d \times 2\frac{d}{g}}$, on which we perform an SVD to get $U_{kv}\Sigma_{kv}B^T_{kv}$. Here, $U_{kv} \in \mathbb{R}^{d \times 2 \frac{d}{g}}$ is a shared subspace for the individual projection matrices, and $\Sigma_{kv}B^T_{kv}$ is discarded. 
We down-project $\Delta X$ by $U_{kv}$, resulting in the same memory footprint as the KV cache, and we quantize and cache this latent distribution. 
Although this approach has the same memory consumption as KV quantization, the deltas are easier to quantize despite the latent projection, giving us higher accuracy for the same bit width (see Section \ref{sec:xquant-cl-ppl}). Crucially, the SVD is performed offline, so there is no additional latency overhead during inference. When computing $\hat{X}_i$, which is the approximation of \textit{X} for layer \textit{i}, we need to merge our accumulator $\hat{X}_{i-1}$ of shape $l \times d$ with the latent $\Delta X_i U_{kv}$. To do this, we up-project $\Delta X_i U_{kv}$ using $U_{kv}^T$, add this result to the accumulator $\hat{X}_{i-1}$, and multiply the updated accumulator $\hat{X}_i$ by $W_{kv}$ to complete the KV cache rematerialization. One concern is whether up-projecting the latent delta by $U_{kv}^T$ is able to retrieve the original delta: for the non-square matrix $W_{kv}$, the SVD produces non-square $U_{kv}$ with orthonormal columns such that $U_{kv}^T U_{kv} = I_{2 \frac{d}{g}}$ but $U_{kv} U_{kv}^T \neq I_{d}$. 
However, when the quantization function $Q$ is the identity function, up-projecting the latent $\Delta X_i U_{kv}$ by $U_{kv}^T$ is a lossless reconstruction of the original $\Delta X_i$ when computing the KV cache for layer $i$:
\begin{align*}
    [K|V]_i &= (\hat{X}_{i-1} + Q(\Delta X_i U_{kv}) \cdot U_{kv}^T) \cdot U_{kv}\Sigma_{kv}B_{kv}^T \\
    &\qquad\text{where} \quad U_{kv}^TU_{kv} = I_{2 \frac{d}{g}},\quad U_{kv}U_{kv}^T \neq I_d, \quad Q(x) = x \\
    &= (\hat{X}_{i-1} U_{kv} + \Delta X_i U_{kv}) \cdot \Sigma_{kv}B_{kv}^T \\
    &= (\hat{X}_{i-1} + \Delta X_i) \cdot U_{kv} \Sigma_{kv}B_{kv}^T \\
    &= (\hat{X}_{i-1} + \Delta X_i) \cdot W_{kv} \\
    &= \hat{X}_{i} \cdot [W_k | W_v] .
\end{align*}

\subsection{System-Level Analysis of Rematerialization}
\label{sec:modeling}

Here, we present system-level modeling that outlines the computational and memory overheads from rematerialization with \OURS and \OURSCL. 
For this analysis, we count a multiply-accumulate operations as two FLOPs.

Assume we are using a model with hidden dimension $d$ and assume sequence length $l$.
To apply \OURS on MHA models, the amount of computation required for rematerialization for a single layer is $2 \cdot 2 \cdot l \cdot d^2$.
Suppose that we apply $e$-bit quantization for $X$.
Then the total number of memory operations in bytes is equal to $\frac{e}{8} \cdot l \cdot d$; whereas for the KV cache quantization, it would have been $2 \cdot \frac{e}{8} \cdot l \cdot d$.
To apply \OURS on GQA models, the amount of computation required for rematerialization for a single layer is $2 \cdot 2 \cdot l \cdot (\frac{d}{g})^2$, where $g$ is the number of query heads that share Keys and Values. 
This is a factor of $g^2$ less floating point operations compared to MHA models.
Suppose that we apply $e$-bit quantization for $X$.
Then the total number of memory operations in bytes is equal to $2 \cdot \frac{e}{8} \cdot l \cdot \frac{d}{g}$, which is the same as KV cache quantization. 
However, for the same $e$-bit quantization, \OURS achieves much higher accuracy than KV cache quantization (see Section \ref{sec:xquant-ppl}). 
Equivalently, \OURS achieves similar accuracy with a smaller $e$ to KV cache quantization with a larger effective $e$, meaning that for the same accuracy, \OURS performs fewer memory operations.

Here, we also provide an example to demonstrate the amount of the rematerialization that can be performed without having the additional compute operations become the latency bottleneck.
Here, we assume the NVIDIA H100 GPU as our hardware target, which has a ridge point of $P = \frac{\mathrm{Peak}\ \mathrm{FLOPs}}{\mathrm{Memory}\ \mathrm{ BW}} = \frac{756\mathrm{TFLOPs}}{2\ \mathrm{TB/s}} = 378$, and we assume that we can overlap the KV cache recomputation with loading the model weights (which corresponds to $2\cdot 12 \cdot d^2$ additional memory operations for a single layer for the Llama-2-7B model).
We can solve for the maximum amount that we can reconstruct without compute becoming the bottleneck:

\begin{equation}
    P = \frac{2 \cdot 2 \cdot l \cdot d^2}{\frac{e}{8} \cdot l \cdot d+ 2\cdot12\cdot d^2}
\end{equation}

Solving this equation for $P=378$, $d=4K$, and $e=2$ gives a maximum sequence length of $2.3K$ that can be rematerialized without having compute operations become the bottleneck.

We can perform similar analysis for the Llama-3.1-8B model (which requires $2\cdot 13 \cdot d^2 + 2\cdot 2 \cdot (\frac{d}{g})^2$ memory operations for the weights for a single layer, including the overhead of loading the $W_k$/$W_v$ matrices in SVD-decomposed form, and which has $g=4$):

\begin{equation}
    P = \frac{2 \cdot 2 \cdot l \cdot (\frac{d}{g})^2}{\frac{e}{8} \cdot l \cdot \frac{d}{g}+ 2\cdot 13\cdot d^2 + 2\cdot 2 \cdot (\frac{d}{g})^2}
\end{equation}

Solving this equation for $P=378$, $d=4K$, $g=4$, and $e=2$ gives a maximum sequence length of $40.6K$ that can be rematerialized without having compute operations become the bottleneck.

With \OURSCL, we perform an additional $2 \cdot l \cdot d$ compute operations for each layer, since we need to add the the cached delta $\Delta \hat{X}_i$ to the accumulator $\hat{X}_{i-1}$. However, this cost is negligible: if we combine it with the aforementioned rematerialization cost, we get $2\cdot2\cdot l\cdot d^2 + 2\cdot l\cdot d = 2\cdot 2\cdot l\cdot d \cdot (d + \frac{1}{2})$. \OURSCL also requires additional memory operations, as we need to load and store both the accumulator and the cached delta at each layer. Since we keep the accumulator in higher precision, we perform $\frac{e_b}{8} \cdot l \cdot d$ memory operations in bytes at each layer, where $e_b$ is the accumulator's precision (typically $e_b=4$ bits; see Section \ref{sec:xquant-cl-ppl}). 
Additionally, we perform $\frac{e}{8} \cdot l \cdot d$ memory operations required to load the cached delta, which is the same as \OURS. 
For GQA models, since we are caching the quantized down-projected $\Delta X_i U_{kv}$ at each layer, the number of memory operations for the delta is $2 \cdot \frac{e}{8} \cdot l \cdot \frac{d}{g}$, which is the same for \OURS applied on GQA models. 
After loading this delta, we up-project it by $U_{kv}^T$ to merge with the accumulator $\hat{X}_{i-1}$, and then apply the Key and Value weight projections onto the updated accumulator. 
In total, this requires $2 \cdot 4 \cdot l \cdot \frac{d}{g} \cdot d$ compute operations.

\section{Empirical Results}
\label{sec:results}

To evaluate our method, we use the Llama-2-7B/13B, Llama-3.1-8B, and Mistral-7B-v0.3 models \cite{touvron2023Llama2,llama31,jiang2023mistral}.
We measure perplexity on the WikiText-2 and C4 datasets \cite{wikitext2,c4}, and we perform downstream task evaluation on the LongBench and GSM8K datasets \cite{bai2024longbench, cobbe2021trainingverifierssolvemath}.
Note that the Llama-2-7B/13B models use MHA, whereas Llama-3.1-8B and Mistral-7B use GQA. 
We compare our method against KIVI \cite{liu2024kivi} and KVQuant \cite{hooper2024kvquant}, two representative state-of-the-art KV cache quantization methods.

For KIVI, we use a stronger baseline (referred to here as KIVI*). The original KIVI method quantizes the Keys after rotary positional embeddings (RoPE) are applied to them \cite{liu2024kivi}. However, \cite{hooper2024kvquant} finds that applying RoPE to the Keys results in a less structured distribution, whereas the Keys have more structured outlier channels pre-RoPE. We therefore follow \cite{hooper2024kvquant} and quantize the Keys before applying RoPE.

For KVQuant, we use the best configuration from \cite{hooper2024kvquant}, which performs non-uniform, per-vector dense-and-sparse quantization. KVQuant's non-uniform quantization requires deriving per-layer sensitivity-weighted non-uniform datatypes from a calibration dataset offline to
better represent the distributions. We follow \cite{hooper2024kvquant} and use 16 calibration samples of sequence length 2K from WikiText-2. 
The dense-and-sparse quantization isolates outliers separately for each vector and preserves them in higher precision.

Note that \OURS and \OURSCL use simple uniform quantization, with no special outlier handling or calibration required. 
For the KV cache quantization baselines, we quantize the Keys per-channel and the Values per-token \cite{liu2024kivi, hooper2024kvquant}.
We use group size of 128 for all quantization experiments. 
For generative tasks, in order to be able to leverage per-channel quantization during decoding, we leave the final residual tokens unquantized (up to group size tokens), similar to the residual method in \cite{liu2024kivi}.
We adopt this residual method across all generation experiments for fair comparison.

\subsection{Main Results}
\label{sec:xquant-ppl}

\begin{table*}[t]
\caption{
\OURS evaluation using perplexity on WikiText-2 and C4.
Left: (MHA) Llama‑2‑7B/13B; Right: (GQA) Llama‑3.1‑8B, and Mistral‑7B.
We provide KV cache size estimates (normalized to the KV cache size of the FP16 baseline) and group rows by similar memory consumption. For MHA models, each group shows \OURS using slightly less memory compared to KIVI*, as \OURS only needs to store scale factors and zero points for a single \textit{X} tensor, whereas KIVI* needs to store the same for \textit{K} and \textit{V}.
}
\label{tab:xquant-ppl}
\vspace{1mm}
\centering
\scriptsize
\renewcommand{\arraystretch}{1.2}
\setlength{\tabcolsep}{6pt}

\begin{minipage}[t]{0.44\textwidth}
\centering
\begin{tabular}{c|c|cc|cc}
\toprule
\multirow{2}{*}{\textbf{Method}} &
\textbf{KV} &
\multicolumn{2}{c|}{\textbf{Llama‑2‑7B}} &
\multicolumn{2}{c}{\textbf{Llama‑2‑13B}} \\
\cmidrule{3-6}
& \textbf{(MHA)} & Wiki2 & C4 & Wiki2 & C4 \\ \midrule\midrule
Baseline          & 1.00 & 5.47 & 7.26 & 4.88 & 6.73 \\
\midrule
KIVI*-4bit     & 0.27 & 5.49 & 7.30 & 4.90 & 6.75 \\
\hc \OURS-8bit       & 0.26 & \textbf{5.47} & \textbf{7.26} & \textbf{4.88} & \textbf{6.73} \\
\midrule
KIVI*-3bit     & 0.20 & 5.58 & 7.40 & 4.97 & 6.83 \\
\midrule
KIVI*-2bit     & 0.14 & 6.42 & 8.46 & 5.61 & 7.67 \\
\hc \OURS-4bit       & 0.13 & \textbf{5.54} & \textbf{7.36} & \textbf{4.94} & \textbf{6.79} \\
\midrule
\hc \OURS-3bit       & 0.10 & 6.65 & 8.65 & 5.22 & 7.07 \\
\bottomrule
\end{tabular}
\end{minipage}
\hfill
\begin{minipage}[t]{0.54\textwidth}
\centering
\begin{tabular}{c|c|cc|cc}
\toprule
\multirow{2}{*}{\textbf{Method}} &
\textbf{KV} &
\multicolumn{2}{c|}{\textbf{Llama‑3.1‑8B}} &
\multicolumn{2}{c}{\textbf{Mistral‑7B}} \\
\cmidrule{3-6}
& \textbf{(GQA)} & Wiki2 & C4 & Wiki2 & C4 \\ \midrule\midrule
Baseline          & 1.00 & 6.24 & 9.54 & 5.32 & 8.47 \\
\midrule
KIVI*-4bit     & 0.27 & 6.31 & 9.66 & 5.34 & 8.51 \\
\hc \OURS-4bit       & 0.27 & \textbf{6.28} & \textbf{9.60} & \textbf{5.33} & \textbf{8.49} \\
\midrule
KIVI*-3bit     & 0.20 & 6.59 & 10.13 & 5.43 & 8.62 \\
\hc \OURS-3bit       & 0.20 & \textbf{6.43} & \textbf{9.89} & \textbf{5.39} & \textbf{8.57} \\
\midrule
KIVI*-2bit     & 0.14 & 9.95 & 15.98 & 6.36 & 9.88 \\
\hc \OURS-2bit       & 0.14 & \textbf{7.74} & \textbf{12.27} & \textbf{5.79} & \textbf{9.13} \\
\bottomrule
\end{tabular}
\end{minipage}
\end{table*}


In Table \ref{tab:xquant-ppl}, we report perplexity results for Llama-2-7B/13B, Llama-3.1-8B, and Mistral-7B models on WikiText-2 and C4. 
Perplexity has been measured using teacher forcing with the output logits of all input tokens. 
For \OURS applied to MHA models, we apply per-token quantization to \textit{X}, whereas for KIVI*, we follow \cite{liu2024kivi, hooper2024kvquant} and we apply per-channel quantization to the pre-RoPE Keys and per-token quantization to the Values. 
For \OURS applied to GQA models, we cache $XU_k$ and $XU_v$, and we find that applying per-channel quantization to the latent $XU_k$ and per-token quantization to the latent $XU_v$ is the best configuration.

Overall, we find that \OURS not only substantially outperforms KIVI* for the same memory footprint, but it also retains accuracy very close to the FP16 baseline for great memory savings. 
For the same memory footprint, \OURS achieves 0.88 and 0.67 less perplexity degradation compared to KIVI* for the Llama-2-7B and Llama-2-13B models, respectively. 
Relative to the FP16 baseline, \OURS achieves under 0.1 perplexity degradation, while using $7.7\times$ less memory. 
Similarly, for GQA models, \OURS pushes the boundaries of 2-bit quantization by achieving up to less than 2.2 perplexity degradation compared to KIVI* on Llama-3.1-8B. 
For Mistral-7B, relative to the FP16 baseline, \OURS achieves $< 0.1$ perplexity degradation in 3-bit with $5\times$ memory savings, and only $0.01$ perplexity degradation in 4-bit with $3.7\times$ memory savings. In 2-bit precision, \OURS achieves 0.57 less perplexity degradation compared to KIVI*, while getting $7.1 \times$ more memory savings relative to the FP16 baseline.

\subsection{Downstream Task Evaluation}

\begin{table*}[h!]
\centering
\caption{
\OURS evaluation on LongBench. We report results for the (MHA) Llama-2-7B-chat and (GQA) Llama-3.1-8B-Instruct models.
We include baseline comparisons with KIVI* \cite{liu2024kivi} (details for this configuration are provided in Section \ref{sec:xquant-ppl}).
We report accuracy on each task as well as average cross-task accuracy.
We also provide KV cache size estimates (normalized to the KV cache size of the FP16 baseline). 
}
\label{tab:longbench}
\footnotesize
\setlength{\tabcolsep}{3.2pt}{
\begin{tabular}{c|c|ccc|ccc|ccc|ccc|cc|c}
\toprule
\multirow{2}{*}{\raisebox{-4.8mm}{\textbf{Config}}} &  \multirow{2}{*}{\raisebox{-4.8mm}{\textbf{KV Budget}}} & \multicolumn{3}{c|}{\textbf{Single-Doc. QA}} & \multicolumn{3}{c|}{\textbf{Multi-Doc. QA}} & \multicolumn{3}{c|}{\textbf{Summarization}} & \multicolumn{3}{c|}{\textbf{Few-shot Learning}} & \multicolumn{2}{c|}{\textbf{Code}} & \multirow{2}{*}{\raisebox{-4.8mm}{\textbf{Avg.}}}\\
\cline{3-16}
& & \rotatebox{45}{\textbf{NQA}} & \rotatebox{45}{\textbf{Qspr}} & \rotatebox{45}{\textbf{MFQA}} & \rotatebox{45}{\textbf{HPQA}} & \rotatebox{45}{\textbf{2Wiki}} & \rotatebox{45}{\textbf{MSQ}} & \rotatebox{45}{\textbf{GRep}} & \rotatebox{45}{\textbf{QMSM  }} & \rotatebox{45}{\textbf{MNews}} & \rotatebox{45}{\textbf{TREC}} & \rotatebox{45}{\textbf{TQA}} & \rotatebox{45}{\textbf{SSum}}  & \rotatebox{45}{\textbf{RB}}& \rotatebox{45}{\textbf{LCC}} & \\ 
\midrule
\multicolumn{17}{c}{\textbf{Llama-2-7B-Chat}} \\ 
\midrule
All KV & 1.00 & 19.4 & 22.1 & 36.7 & 27.8 & 31.5 & 8.3 & 26.9 & 20.7 & 26.2 & 64.0 & 83.3 & 41.3 & 52.4 & 58.3 & 37.1 \\
\midrule
KIVI*-4bit & 0.27 & 18.7 & 21.1 & 37.3 & 28.7 & 31.9 & 8.4 & 27.3 & 21.4 & 25.9 & 64.0 & 83.0 & 40.9 & 52.6 & 57.5 & 37.1 \\
KIVI*-3bit & 0.20 & 18.1 & 21.2 & 37.3 & 26.0 & 30.4 & 8.7 & 26.7 & 21.2 & 26.2 & 61.0 & 84.1 & 40.7 & 52.0 & 57.2 & 36.5 \\
\midrule
KIVI*-2bit & 0.14 & 13.4 & 21.0 & 31.1 & 19.8 & 23.1 & 4.9 & 24.1 & 20.7 & 25.2 & 55.5 & 70.6 & 39.6 & 48.4 & 51.6 & 32.1 \\
\hc \OURS-4bit & 0.13 & 19.0 & 21.3 & 36.4 & 27.7 & 32.6 & 8.6 & 26.8 & 21.5 & 26.2 & 66.5 & 82.4 & 41.1 & 54.8 & 60.5 & \textbf{37.5} \\
\midrule
\hc \OURS-3bit & 0.10 & 11.9 & 16.2 & 35.3 & 21.4 & 19.0 & 7.3 & 27.1 & 20.5 & 25.2 & 57.5 & 77.0 & 38.1 & 49.7 & 52.4 & 32.8 \\
\midrule
\multicolumn{17}{c}{\textbf{Llama-3.1-8B-Instruct}}  \\ 
\midrule
All KV & 1.00 & 31.2 & 45.5 & 53.8 & 55.0 & 47.1 & 31.4 & 34.8 & 25.3 & 27.5 & 72.5 & 91.7 & 43.7 & 56.6 & 63.4 & 48.5 \\
\midrule
KIVI*-4bit& 0.27  & 30.0 & 46.0 & 54.4 & 55.7 & 45.5 & 31.0 & 34.6 & 25.3 & 27.2 & 74.0 & 90.5 & 44.1 & 55.8 & 64.0 & 48.4 \\
\hc \OURS-4bit & 0.27 & 30.8 & 46.7 & 55.1 & 54.5 & 47.9 & 30.5 & 34.9 & 25.7 & 27.5 & 73.0 & 91.1 & 43.6 & 55.9 & 62.6 & \textbf{48.6} \\
\midrule
KIVI*-3bit & 0.20  & 29.6 & 45.8 & 55.4 & 54.0 & 43.0 & 31.4 & 34.3 & 25.2 & 27.4 & 70.5 & 90.7 & 43.5 & 49.0 & 61.6 & 47.2 \\
\hc \OURS-3bit & 0.20 & 30.9 & 46.3 & 52.0 & 54.4 & 43.9 & 30.5 & 34.6 & 25.1 & 27.4 & 73.5 & 91.6 & 43.4 & 53.9 & 62.7 & \textbf{47.9} \\
\midrule
KIVI*-2bit & 0.14 & 23.7 & 36.8 & 41.1 & 44.4 & 30.0 & 21.9 & 30.5 & 24.4 & 26.0 & 68.0 & 87.8 & 44.7 & 38.2 & 41.0 & 39.9 \\
\hc \OURS-2bit & 0.14 & 25.3 & 38.3 & 49.6 & 48.6 & 35.5 & 28.4 & 34.0 & 24.8 & 27.3 & 69.0 & 88.7 & 44.7 & 52.3 & 57.6 & \textbf{44.6} \\
\bottomrule
\end{tabular}
}
\end{table*}

\begin{table}[t]
\caption{
\OURS evaluation on GSM8K. We report results for the Llama-2-7B-chat model.
We include baseline comparisons with KIVI* \cite{hooper2024kvquant} (details for this configuration are provided in Section \ref{sec:xquant-ppl}).
We also provide KV cache size estimates (normalized to the KV cache size of the FP16 baseline).
}\label{tab:gsm8k}
\centering
\footnotesize{
\setlength{\tabcolsep}{5.5pt}{
\begin{tabular}{c|c|c}
        \toprule
\textbf{Config} & \textbf{Accuracy} & \textbf{KV Cache Size (Normalized)} \\
\midrule
All KV & 0.132 & 1.00  \\
\midrule
KIVI*-4bit & 0.149 & 0.27  \\
KIVI*-3bit & 0.130 & 0.20  \\
\midrule
KIVI*-2bit & 0.092 & 0.14 \\
\hc \OURS-4bit & \textbf{0.129} & 0.13 \\
\midrule
\hc \OURS-3bit & 0.086 & 0.10 \\
\bottomrule
\end{tabular}
}
}
\end{table}


We also provide cross-task evaluation to demonstrate the applicability of our strategy for a range of downstream tasks.
Table \ref{tab:longbench} provides evaluation on LongBench, a benchmark suite containing a range of long-context length tasks including in-context learning, document Q/A, summarization, and coding tasks. 
We report \OURS results for the Llama-2-7B-Chat (MHA) and Llama-3.1-8B (GQA) models, and we also provide baseline comparisons against KIVI* \cite{hooper2024kvquant}. 
For Llama-2-7B-Chat, \OURS-4bit attains the same accuracy as KIVI*-4bit and the same accuracy as the baseline, while providing 2$\times$ additional memory compression.
For Llama-3.1-8B-Instruct, \OURS provides improved accuracy on average for the same bit width across all precision settings.

We also include results on long-generation reasoning tasks to demonstrate the applicability of our method for complex reasoning.
Table \ref{tab:gsm8k} provides evaluation for the GSM8K dataset \cite{cobbe2021trainingverifierssolvemath} (using lm-eval-harness \cite{lm-eval-harness}), which evaluates arithmetic reasoning capabilities.
We use the chain-of-thought (CoT) configuration, and we report strict match accuracy.
We report results for the Llama-2-7B-Chat model, and we also provide baseline comparisons against KIVI* \cite{hooper2024kvquant}.
We find that \OURS-4bit attains similar accuracy to KIVI*-3bit, while providing 1.5$\times$ additional memory savings, and outperforms KIVI*-2bit while using $2\times$ less memory.

\subsection{Results with Cross-Layer Compression Method}
\label{sec:xquant-cl-ppl}

\begin{table*}[t]
\caption{\OURS evaluation using perplexity (PPL) on WikiText-2 and C4 for (MHA) Llama‑2‑7B/13B and (GQA) Llama‑3.1‑8B and Mistral‑7B models.
We provide KV cache size estimates (normalized to the KV cache size of the FP16 baseline). Note that the first 3 layers are quantized in 4-bit for KIVI*, \OURS, and \OURSCL. This is done so that these methods have a comparable memory footprint to KVQuant, which stores additional memory for outliers.}
\label{tab:xquant-cl-ppl}
\centering
\footnotesize
\renewcommand{\arraystretch}{1.2}
\setlength{\tabcolsep}{6pt}
\begin{tabular}{c|c|cc|cc||c|cc|cc}
\toprule
\multirow{3}{*}{\textbf{Method}} & 
\multirow{2}{*}{\textbf{KV}} &
\multicolumn{2}{c|}{\textbf{Llama‑2‑7B}} &
\multicolumn{2}{c||}{\textbf{Llama‑2‑13B}} &
\multirow{2}{*}{\textbf{KV}} &
\multicolumn{2}{c|}{\textbf{Llama‑3.1‑8B}} &
\multicolumn{2}{c}{\textbf{Mistral‑7B}} \\
&  & Wiki2 & C4 & Wiki2 & C4 &  & Wiki2 & C4 & Wiki2 & C4 \\ 
& \multicolumn{5}{c||}{\textbf{(MHA)}} & \multicolumn{5}{c}{\textbf{(GQA)}} \\ 
\midrule\midrule
baseline
    & 1.00 & 5.47 & 7.26 & 4.88 & 6.73
    & 1.00 & 6.24 & 9.54 & 5.32 & 8.48 \\
\midrule
KIVI*-4bit
    & 0.27 & 5.49 & 7.30 & 4.90 & 6.75
    & 0.27 & 6.30 & 9.65 & 5.34 & 8.50 \\
KVQuant-4bit-1\%
    & 0.27 & 5.49 & 7.28 & 4.90 & 6.75
    & 0.27 & 6.30 & 9.63 & 5.34 & 8.50 \\
\hc
\OURS-4bit
    & 0.13 & 5.54 & 7.36 & 4.94 & 6.79
    & 0.27 & 6.31 & 9.65 & 5.33 & 8.50 \\
\hd
\OURSCL-4bit
    & \textbf{0.13} & \textbf{5.48} & \textbf{7.27} & \textbf{4.89} & \textbf{6.74}
    & \textbf{0.27} & \textbf{6.29} & \textbf{9.61} & \textbf{5.32} & \textbf{8.49} \\
\midrule
KIVI*-3bit
    & 0.21 & 5.57 & 7.39 & 4.96 & 6.82
    & 0.21 & 6.55 & 10.05 & 5.41 & 8.61 \\
KVQuant-3bit-1\%
    & 0.21 & 5.56 & 7.36 & 4.96 & 6.81
    & 0.21 & 6.48 & 9.90 & 5.41 & 8.58 \\
\hc
\OURS-3bit
    & 0.10 & 6.10 & 8.20 & 5.10 & 6.95
    & 0.21 & 6.43 & 9.87 & 5.37 & 8.57 \\
\hd
\OURSCL-3bit
    & \textbf{0.10} & \textbf{5.48} & \textbf{7.28} & \textbf{4.92} & \textbf{6.78}
    & \textbf{0.21} & \textbf{6.32} & \textbf{9.67} & \textbf{5.34} & \textbf{8.51} \\
\midrule
KIVI*-2bit
    & 0.15 & 6.20 & 8.22 & 5.46 & 7.49
    & 0.15 & 8.72 & 13.54 & 6.14 & 9.59 \\
KVQuant-2bit-1\%
    & 0.15 & 5.99 & 7.83 & 5.34 & 7.23
    & 0.15 & 7.45 & 11.49 & 5.87 & 9.10 \\
\hc
\OURS-2bit
    & 0.08 & 11.21 & 15.27 & 7.00 & 10.25
    & 0.15 & 7.38 & 11.54 & 5.69 & 9.01 \\
\hd
\OURSCL-2bit
    & \textbf{0.08} & \textbf{5.57} & \textbf{7.39} & \textbf{5.11} & \textbf{7.09}
    & \textbf{0.15} & \textbf{6.60} & \textbf{10.15} & \textbf{5.46} & \textbf{8.67} \\
\bottomrule
\end{tabular}
\end{table*}


In Table \ref{tab:xquant-cl-ppl}, we provide an evaluation for our \OURSCL method, which retains near FP16 accuracy while achieving great memory savings with standard asymmetric uniform quantization. We report results for Llama-2-7B/13, Llama-3.1-8B, and Mistral-7B-v0.3. We also provide baseline comparisons against KIVI* \cite{liu2024kivi}, which also uses asymmetric uniform quantization, and KVQuant \cite{hooper2024kvquant}, which uses non-uniform, per-vector dense-and-sparse quantization (see Section \ref{sec:xquant-ppl} for all configuration details). In our results, we list KVQuant as KVQuant-$\langle e\rangle$bit-$\langle o \rangle\%$, where $\langle e\rangle$ is the quantization bit-width and $\langle o \rangle\%$ is the percentage of outliers per layer stored in a sparse, high precision format. We use an outlier threshold of $1\%$, which is the best configuration shown in \cite{hooper2024kvquant}. Note that for KIVI*, \OURS, and \OURSCL, we keep the first 3 layers in higher precision (4-bit). We find that keeping these early layers in higher precision noticeably mitigates perplexity degradation and also results in a comparable memory footprint with KVQuant, which uses additional memory to keep outliers in higher precision. This is in keeping with the findings of \cite{jastrzebski2018residual} which showed that the first few layers of a network with residual connections do substantial representation learning (large transformations of the input) whereas later layers apply small iterative refinements to the input. Thus, the deltas for the first few layers are harder to quantize, whereas the deltas for the remaining layers of the network are much easier to quantize in low-bit precision. For \OURSCL, we use the third layer as the higher-precision base layer which becomes the accumulator.

For Llama-2-7B, relative to the FP16 baseline, \OURSCL results in only $0.1$ perplexity degradation in 2-bit with $12.5\times$ memory savings, and only $0.01$ perplexity degradation in 3-bit with $10\times$ memory savings. Compared to KVQuant-2bit-1\%, \OURSCL-2bit results in $0.42$ less perplexity degradation while using $1.9\times$ less memory. Similarly for Llama-2-13B, \OURSCL-2bit achieves 0.23 less perplexity degradation compared to KVQuant-2bit-1\% on WikiText-2 while using $1.9\times$ less memory. On Llama-3.1-8B, in 2-bit precision, \OURSCL retains a perplexity of 10.15 on C4, which is more than 3 points less than KIVI*-2bit (13.54) and more than 1 point less than KVQuant-2bit-1\% (11.49). On Mistral-7B, \OURSCL-2bit saves $6.7\times$ memory while only facing a perplexity degradation of 0.14 compared to the FP16 baseline on WikiText-2. Impressively, for only a perplexity degradation of 0.12 relative to KVQuant-4bit-1\% on WikiText-2, \OURSCL-2bit uses $1.8\times$ less memory than KVQuant-4bit-1\%. Overall, \OURSCL outperforms state-of-the-art KV cache quantization methods for ultra low precision bit widths, achieving near FP16 accuracy with $1.5-2\times$ memory savings over KV cache quantization and $6-12\times$ memory savings over the FP16 baseline.

\section{Conclusion}

As compute capabilities continue to outpace memory bandwidth and capacity on modern GPU hardware, LLM inference is increasingly becoming more memory-bandwidth bound. 
In light of this hardware scaling trend, a natural strategy is to perform additional compute operations in order to reduce memory requirements. 
In this work, we adopt a forward-looking vision to speed up inference of LLMs by exploiting the compute scaling trends on newer generation hardware. 
Specifically, LLM inference is typically memory-bandwidth bound due to loading the large KV cache when generate each token. 
To address this, we aim to reduce the memory requirements for LLM inference through reducing the size of the KV cache activations in exchange for higher computation cost, thus reducing the number of memory operations needed to generate each token and speeding up inference.
We propose \OURS, which quantizes the layer input activations in order to reduce memory consumption by 2$\times$ relative to KV caching, and rematerializes the Keys and Values on-the-fly during inference.
We then extend our basic method and propose \OURSCL, which exploits the cross-layer compressibility of the \textit{X} embeddings between successive layers. 
We find that using simple uniform quantization, \OURS and \OURSCL surpass state-of-the-art KV cache quantization methods like KVQuant that use non-uniform quantization and dense-and-sparse quantization, while also retaining accuracy close to the FP16 baseline. 
Relative to the FP16 baseline, \OURS achieves under 0.1 perplexity degradation while also using $7.7\times$ less memory for the Llama-2-7B and Llama-2-13B models. \OURSCL achieves $12.5\times$ memory savings with only 0.1 perplexity degradation in 2-bit precision, and $10\times$ memory savings with only 0.01 perplexity degradation in 3-bit precision compared to the FP16 baseline. Both \OURS and \OURSCL reduce perplexity degradation by several points compared to KIVI* and KVQuant low-bit precision quantization.
With the growing discrepancy between compute and memory capabilities on hardware platforms, rematerialization methods like \OURS can help exploit the available computation in order to accelerate memory bandwidth-bound LLM inference, while also retaining near FP16 accuracy even in low-bit precision.

\subsection*{Limitations}

Our work focuses on reducing the memory requirements for LLM inference by quantizing the input \textit{X} embeddings and rematerializing KV cache activations.
While this approach allows for aggressive memory compression with minimal accuracy loss, it requires additional compute operations to perform rematerialization, which may increase latency on particular hardware platforms.
Additionally, \OURSCL reduces the memory capacity requirements, but it requires additional compute and memory operations in order to rematerialize KV cache activations due to having to load the accumulator. However, in memory-constrained scenarios where the goal is to attain near FP16 accuracy, \OURSCL can be an optimal choice.

\subsection*{Acknowledgements}
We acknowledge gracious support from the FuriosaAI team including Jihoon Yoon, Kevin Galim, Heeju Kim, and Hyung Il Koo, as well as from Intel, Apple, NVIDIA, and Mozilla.
We also appreciate the support from Microsoft through their Accelerating Foundation Model Research.
Furthermore, we appreciate support from
Google Cloud, the Google TRC team and Prof. David Patterson.
Prof. Keutzer's lab is sponsored by the Intel corporation, UC Berkeley oneAPI Center of Excellence, Intel VLAB team, as well as funding through BDD and BAIR.
MWM would also like to acknowledge DARPA, DOE, NSF, and ONR.
This work was supported in part by the Director, Office of Science, Office of Advanced Scientific Computing Research, of the U.S. Department of Energy under Contract No. DE-AC02-05CH11231.
Our conclusions do not necessarily reflect the position or the policy of our sponsors, and no official endorsement should be~inferred.

\bibliographystyle{plain}
\bibliography{references}

\begin{thebibliography}{10}

\bibitem{ainslie2023gqa}
Joshua Ainslie, James Lee-Thorp, Michiel de~Jong, Yury Zemlyanskiy, Federico Lebron, and Sumit Sanghai.
\newblock Gqa: Training generalized multi-query transformer models from multi-head checkpoints.
\newblock In {\em Proceedings of the 2023 Conference on Empirical Methods in Natural Language Processing}, pages 4895--4901, 2023.

\bibitem{bai2024longbench}
Yushi Bai, Xin Lv, Jiajie Zhang, Hongchang Lyu, Jiankai Tang, Zhidian Huang, Zhengxiao Du, Xiao Liu, Aohan Zeng, Lei Hou, et~al.
\newblock Longbench: A bilingual, multitask benchmark for long context understanding.
\newblock In {\em Proceedings of the 62nd Annual Meeting of the Association for Computational Linguistics (Volume 1: Long Papers)}, pages 3119--3137, 2024.

\bibitem{brown2020language}
Tom Brown, Benjamin Mann, Nick Ryder, Melanie Subbiah, Jared~D Kaplan, Prafulla Dhariwal, Arvind Neelakantan, Pranav Shyam, Girish Sastry, Amanda Askell, et~al.
\newblock Language models are few-shot learners.
\newblock {\em Advances in neural information processing systems}, 33:1877--1901, 2020.

\bibitem{chang2025xkv}
Chi-Chih Chang, Chien-Yu Lin, Yash Akhauri, Wei-Cheng Lin, Kai-Chiang Wu, Luis Ceze, and Mohamed~S Abdelfattah.
\newblock xkv: Cross-layer svd for kv-cache compression.
\newblock {\em arXiv preprint arXiv:2503.18893}, 2025.

\bibitem{chang2024palu}
Chi-Chih Chang, Wei-Cheng Lin, Chien-Yu Lin, Chong-Yan Chen, Yu-Fang Hu, Pei-Shuo Wang, Ning-Chi Huang, Luis Ceze, and Kai-Chiang Wu.
\newblock Palu: Compressing kv-cache with low-rank projection.
\newblock In {\em Proceedings of International Conference on Learning Representations (ICLR)}, April 2025.

\bibitem{chowdhery2023palm}
Aakanksha Chowdhery, Sharan Narang, Jacob Devlin, Maarten Bosma, Gaurav Mishra, Adam Roberts, Paul Barham, Hyung~Won Chung, Charles Sutton, Sebastian Gehrmann, et~al.
\newblock Palm: Scaling language modeling with pathways.
\newblock {\em Journal of Machine Learning Research}, 24(240):1--113, 2023.

\bibitem{cobbe2021trainingverifierssolvemath}
Karl Cobbe, Vineet Kosaraju, Mohammad Bavarian, Mark Chen, Heewoo Jun, Lukasz Kaiser, Matthias Plappert, Jerry Tworek, Jacob Hilton, Reiichiro Nakano, Christopher Hesse, and John Schulman.
\newblock Training verifiers to solve math word problems, 2021.

\bibitem{dubey2024llama}
Abhimanyu Dubey, Abhinav Jauhri, Abhinav Pandey, Abhishek Kadian, Ahmad Al-Dahle, Aiesha Letman, Akhil Mathur, Alan Schelten, Amy Yang, Angela Fan, et~al.
\newblock The llama 3 herd of models.
\newblock {\em arXiv preprint arXiv:2407.21783}, 2024.

\bibitem{lm-eval-harness}
Leo Gao, Jonathan Tow, Baber Abbasi, Stella Biderman, Sid Black, Anthony DiPofi, Charles Foster, Laurence Golding, Jeffrey Hsu, Alain Le~Noac'h, Haonan Li, Kyle McDonell, Niklas Muennighoff, Chris Ociepa, Jason Phang, Laria Reynolds, Hailey Schoelkopf, Aviya Skowron, Lintang Sutawika, Eric Tang, Anish Thite, Ben Wang, Kevin Wang, and Andy Zou.
\newblock The language model evaluation harness, 07 2024.

\bibitem{gao2025fast}
Shiwei Gao, Youmin Chen, and Jiwu Shu.
\newblock Fast state restoration in llm serving with hcache.
\newblock In {\em Proceedings of the Twentieth European Conference on Computer Systems}, pages 128--143, 2025.

\bibitem{gholami2024ai}
Amir Gholami, Zhewei Yao, Sehoon Kim, Coleman Hooper, Michael~W Mahoney, and Kurt Keutzer.
\newblock Ai and memory wall.
\newblock {\em IEEE Micro}, 2024.

\bibitem{graef2025slim}
Nils Graef and Andrew Wasielewski.
\newblock Slim attention: cut your context memory in half without loss--k-cache is all you need for mha.
\newblock {\em arXiv preprint arXiv:2503.05840}, 2025.

\bibitem{han2025polarquant}
Insu Han, Praneeth Kacham, Amin Karbasi, Vahab Mirrokni, and Amir Zandieh.
\newblock Polarquant: Quantizing kv caches with polar transformation.
\newblock {\em arXiv preprint arXiv:2502.02617}, 2025.

\bibitem{he2024zipcache}
Yefei He, Luoming Zhang, Weijia Wu, Jing Liu, Hong Zhou, and Bohan Zhuang.
\newblock Zipcache: Accurate and efficient kv cache quantization with salient token identification.
\newblock {\em Advances in Neural Information Processing Systems}, 37:68287--68307, 2024.

\bibitem{hooper2024kvquant}
Coleman Hooper, Sehoon Kim, Hiva Mohammadzadeh, Michael~W Mahoney, Yakun~S Shao, Kurt Keutzer, and Amir Gholami.
\newblock Kvquant: Towards 10 million context length llm inference with kv cache quantization.
\newblock {\em Advances in Neural Information Processing Systems}, 37:1270--1303, 2024.

\bibitem{jain2020checkmate}
Paras Jain, Ajay Jain, Aniruddha Nrusimha, Amir Gholami, Pieter Abbeel, Joseph Gonzalez, Kurt Keutzer, and Ion Stoica.
\newblock Checkmate: Breaking the memory wall with optimal tensor rematerialization.
\newblock {\em Proceedings of Machine Learning and Systems}, 2:497--511, 2020.

\bibitem{jastrzebski2018residual}
Stanisław Jastrzebski, Devansh Arpit, Nicolas Ballas, Vikas Verma, Tong Che, and Yoshua Bengio.
\newblock Residual connections encourage iterative inference.
\newblock In {\em International Conference on Learning Representations}, 2018.

\bibitem{jiang2023mistral}
Albert~Q Jiang, Alexandre Sablayrolles, Arthur Mensch, Chris Bamford, Devendra~Singh Chaplot, Diego de~las Casas, Florian Bressand, Gianna Lengyel, Guillaume Lample, Lucile Saulnier, et~al.
\newblock Mistral 7b.
\newblock {\em arXiv preprint arXiv:2310.06825}, 2023.

\bibitem{kim2023full}
Sehoon Kim, Coleman Hooper, Thanakul Wattanawong, Minwoo Kang, Ruohan Yan, Hasan Genc, Grace Dinh, Qijing Huang, Kurt Keutzer, Michael~W. Mahoney, Sophia Shao, and Amir Gholami.
\newblock Full stack optimization of transformer inference.
\newblock In {\em Architecture and System Support for Transformer Models (ASSYST @ISCA 2023)}, 2023.

\bibitem{kim2024squeezellm}
Sehoon Kim, Coleman Richard~Charles Hooper, Amir Gholami, Zhen Dong, Xiuyu Li, Sheng Shen, Michael~W Mahoney, and Kurt Keutzer.
\newblock Squeezellm: Dense-and-sparse quantization.
\newblock In {\em International Conference on Machine Learning}, pages 23901--23923. PMLR, 2024.

\bibitem{lee2025efficient}
Sanghyeon Lee, Hongbeen Kim, Soojin Hwang, Guseul Heo, Minwoo Noh, and Jaehyuk Huh.
\newblock Efficient llm inference with activation checkpointing and hybrid caching.
\newblock {\em arXiv preprint arXiv:2501.01792}, 2025.

\bibitem{liu2024intactkv}
Ruikang Liu, Haoli Bai, Haokun Lin, Yuening Li, Han Gao, Zhengzhuo Xu, Lu~Hou, Jun Yao, and Chun Yuan.
\newblock Intactkv: Improving large language model quantization by keeping pivot tokens intact.
\newblock In {\em Findings of the Association for Computational Linguistics ACL 2024}, pages 7716--7741, 2024.

\bibitem{liu2024kivi}
Zirui Liu, Jiayi Yuan, Hongye Jin, Shaochen Zhong, Zhaozhuo Xu, Vladimir Braverman, Beidi Chen, and Xia Hu.
\newblock Kivi: A tuning-free asymmetric 2bit quantization for kv cache.
\newblock In {\em International Conference on Machine Learning}, pages 32332--32344. PMLR, 2024.

\bibitem{wikitext2}
Stephen Merity, Caiming Xiong, James Bradbury, and Richard Socher.
\newblock Pointer sentinel mixture models, 2016.

\bibitem{llama31}
Meta.
\newblock Llama 3.1: \url{{https://ai.meta.com/blog/meta-llama-3-1}}, 2024.

\bibitem{c4}
Colin Raffel, Noam Shazeer, Adam Roberts, Katherine Lee, Sharan Narang, Michael Matena, Yanqi Zhou, Wei Li, and Peter~J. Liu.
\newblock Exploring the limits of transfer learning with a unified text-to-text transformer.
\newblock {\em arXiv e-prints}, 2019.

\bibitem{sadhukhan2025magicdecbreakinglatencythroughputtradeoff}
Ranajoy Sadhukhan, Jian Chen, Zhuoming Chen, Vashisth Tiwari, Ruihang Lai, Jinyuan Shi, Ian En-Hsu Yen, Avner May, Tianqi Chen, and Beidi Chen.
\newblock Magicdec: Breaking the latency-throughput tradeoff for long context generation with speculative decoding.
\newblock In {\em The Thirteenth International Conference on Learning Representations}.

\bibitem{saxena2024eigen}
Utkarsh Saxena, Gobinda Saha, Sakshi Choudhary, and Kaushik Roy.
\newblock Eigen attention: Attention in low-rank space for kv cache compression.
\newblock In {\em Findings of the Association for Computational Linguistics: EMNLP 2024}, pages 15332--15344, 2024.

\bibitem{singhania2024loki}
Prajwal Singhania, Siddharth Singh, Shwai He, Soheil Feizi, and Abhinav Bhatele.
\newblock Loki: Low-rank keys for efficient sparse attention.
\newblock {\em Advances in Neural Information Processing Systems}, 37:16692--16723, 2024.

\bibitem{su2024roformer}
Jianlin Su, Murtadha Ahmed, Yu~Lu, Shengfeng Pan, Wen Bo, and Yunfeng Liu.
\newblock Roformer: Enhanced transformer with rotary position embedding.
\newblock {\em Neurocomputing}, 568:127063, 2024.

\bibitem{team2023gemini}
Gemini Team, Rohan Anil, Sebastian Borgeaud, Jean-Baptiste Alayrac, Jiahui Yu, Radu Soricut, Johan Schalkwyk, Andrew~M Dai, Anja Hauth, Katie Millican, et~al.
\newblock Gemini: a family of highly capable multimodal models.
\newblock {\em arXiv preprint arXiv:2312.11805}, 2023.

\bibitem{gemmateam2025gemma3technicalreport}
Gemma Team, Aishwarya Kamath, Johan Ferret, Shreya Pathak, Nino Vieillard, Ramona Merhej, Sarah Perrin, Tatiana Matejovicova, Alexandre Ramé, Morgane Rivière, Louis Rouillard, Thomas Mesnard, Geoffrey Cideron, Jean bastien Grill, Sabela Ramos, Edouard Yvinec, Michelle Casbon, Etienne Pot, Ivo Penchev, Gaël Liu, Francesco Visin, Kathleen Kenealy, Lucas Beyer, Xiaohai Zhai, Anton Tsitsulin, Robert Busa-Fekete, Alex Feng, Noveen Sachdeva, Benjamin Coleman, Yi~Gao, Basil Mustafa, Iain Barr, Emilio Parisotto, David Tian, Matan Eyal, Colin Cherry, Jan-Thorsten Peter, Danila Sinopalnikov, Surya Bhupatiraju, Rishabh Agarwal, Mehran Kazemi, Dan Malkin, Ravin Kumar, David Vilar, Idan Brusilovsky, Jiaming Luo, Andreas Steiner, Abe Friesen, Abhanshu Sharma, Abheesht Sharma, Adi~Mayrav Gilady, Adrian Goedeckemeyer, Alaa Saade, Alex Feng, Alexander Kolesnikov, Alexei Bendebury, Alvin Abdagic, Amit Vadi, András György, André~Susano Pinto, Anil Das, Ankur Bapna, Antoine Miech, Antoine Yang, Antonia Paterson, Ashish
  Shenoy, Ayan Chakrabarti, Bilal Piot, Bo~Wu, Bobak Shahriari, Bryce Petrini, Charlie Chen, Charline~Le Lan, Christopher~A. Choquette-Choo, CJ~Carey, Cormac Brick, Daniel Deutsch, Danielle Eisenbud, Dee Cattle, Derek Cheng, Dimitris Paparas, Divyashree~Shivakumar Sreepathihalli, Doug Reid, Dustin Tran, Dustin Zelle, Eric Noland, Erwin Huizenga, Eugene Kharitonov, Frederick Liu, Gagik Amirkhanyan, Glenn Cameron, Hadi Hashemi, Hanna Klimczak-Plucińska, Harman Singh, Harsh Mehta, Harshal~Tushar Lehri, Hussein Hazimeh, Ian Ballantyne, Idan Szpektor, Ivan Nardini, Jean Pouget-Abadie, Jetha Chan, Joe Stanton, John Wieting, Jonathan Lai, Jordi Orbay, Joseph Fernandez, Josh Newlan, Ju~yeong Ji, Jyotinder Singh, Kat Black, Kathy Yu, Kevin Hui, Kiran Vodrahalli, Klaus Greff, Linhai Qiu, Marcella Valentine, Marina Coelho, Marvin Ritter, Matt Hoffman, Matthew Watson, Mayank Chaturvedi, Michael Moynihan, Min Ma, Nabila Babar, Natasha Noy, Nathan Byrd, Nick Roy, Nikola Momchev, Nilay Chauhan, Noveen Sachdeva, Oskar
  Bunyan, Pankil Botarda, Paul Caron, Paul~Kishan Rubenstein, Phil Culliton, Philipp Schmid, Pier~Giuseppe Sessa, Pingmei Xu, Piotr Stanczyk, Pouya Tafti, Rakesh Shivanna, Renjie Wu, Renke Pan, Reza Rokni, Rob Willoughby, Rohith Vallu, Ryan Mullins, Sammy Jerome, Sara Smoot, Sertan Girgin, Shariq Iqbal, Shashir Reddy, Shruti Sheth, Siim Põder, Sijal Bhatnagar, Sindhu~Raghuram Panyam, Sivan Eiger, Susan Zhang, Tianqi Liu, Trevor Yacovone, Tyler Liechty, Uday Kalra, Utku Evci, Vedant Misra, Vincent Roseberry, Vlad Feinberg, Vlad Kolesnikov, Woohyun Han, Woosuk Kwon, Xi~Chen, Yinlam Chow, Yuvein Zhu, Zichuan Wei, Zoltan Egyed, Victor Cotruta, Minh Giang, Phoebe Kirk, Anand Rao, Kat Black, Nabila Babar, Jessica Lo, Erica Moreira, Luiz~Gustavo Martins, Omar Sanseviero, Lucas Gonzalez, Zach Gleicher, Tris Warkentin, Vahab Mirrokni, Evan Senter, Eli Collins, Joelle Barral, Zoubin Ghahramani, Raia Hadsell, Yossi Matias, D.~Sculley, Slav Petrov, Noah Fiedel, Noam Shazeer, Oriol Vinyals, Jeff Dean, Demis Hassabis,
  Koray Kavukcuoglu, Clement Farabet, Elena Buchatskaya, Jean-Baptiste Alayrac, Rohan Anil, Dmitry, Lepikhin, Sebastian Borgeaud, Olivier Bachem, Armand Joulin, Alek Andreev, Cassidy Hardin, Robert Dadashi, and Léonard Hussenot.
\newblock Gemma 3 technical report, 2025.

\bibitem{tiwari2025quantspec}
Rishabh Tiwari, Haocheng Xi, Aditya Tomar, Coleman Hooper, Sehoon Kim, Maxwell Horton, Mahyar Najibi, Michael~W Mahoney, Kurt Keutzer, and Amir Gholami.
\newblock Quantspec: Self-speculative decoding with hierarchical quantized kv cache.
\newblock In {\em International Conference on Machine Learning}, 2025.

\bibitem{touvron2023llama}
Hugo Touvron, Thibaut Lavril, Gautier Izacard, Xavier Martinet, Marie-Anne Lachaux, Timoth{\'e}e Lacroix, Baptiste Rozi{\`e}re, Naman Goyal, Eric Hambro, Faisal Azhar, et~al.
\newblock {LLaMA}: Open and efficient foundation language models.
\newblock {\em arXiv preprint arXiv:2302.13971}, 2023.

\bibitem{touvron2023Llama2}
Hugo Touvron, Louis Martin, Kevin Stone, Peter Albert, Amjad Almahairi, Yasmine Babaei, Nikolay Bashlykov, Soumya Batra, Prajjwal Bhargava, Shruti Bhosale, et~al.
\newblock Llama 2: Open foundation and fine-tuned chat models.
\newblock {\em arXiv preprint arXiv:2307.09288}, 2023.

\bibitem{vaswani2017attention}
Ashish Vaswani, Noam Shazeer, Niki Parmar, Jakob Uszkoreit, Llion Jones, Aidan~N Gomez, {\L}ukasz Kaiser, and Illia Polosukhin.
\newblock Attention is all you need.
\newblock In {\em Advances in neural information processing systems}, pages 5998--6008, 2017.

\bibitem{williams2009roofline}
Samuel Williams, Andrew Waterman, and David Patterson.
\newblock Roofline: an insightful visual performance model for multicore architectures.
\newblock {\em Communications of the ACM}, 52(4):65--76, 2009.

\bibitem{wu2025polarquant}
Songhao Wu, Ang Lv, Xiao Feng, Yufei Zhang, Xun Zhang, Guojun Yin, Wei Lin, and Rui Yan.
\newblock Polarquant: Leveraging polar transformation for efficient key cache quantization and decoding acceleration.
\newblock {\em arXiv preprint arXiv:2502.00527}, 2025.

\bibitem{xiao2023efficient}
Guangxuan Xiao, Yuandong Tian, Beidi Chen, Song Han, and Mike Lewis.
\newblock Efficient streaming language models with attention sinks.
\newblock In {\em The Twelfth International Conference on Learning Representations}.

\bibitem{xiong2020layernormalizationtransformerarchitecture}
Ruibin Xiong, Yunchang Yang, Di~He, Kai Zheng, Shuxin Zheng, Chen Xing, Huishuai Zhang, Yanyan Lan, Liwei Wang, and Tieyan Liu.
\newblock On layer normalization in the transformer architecture.
\newblock In {\em International conference on machine learning}, pages 10524--10533. PMLR, 2020.

\bibitem{yan2025recalkv}
Xianglong Yan, Zhiteng Li, Tianao Zhang, Linghe Kong, Yulun Zhang, and Xiaokang Yang.
\newblock Recalkv: Low-rank kv cache compression via head reordering and offline calibration.
\newblock {\em arXiv preprint arXiv:2505.24357}, 2025.

\bibitem{yan2021attention}
Yu~Yan, Jiusheng Chen, Weizhen Qi, Nikhil Bhendawade, Yeyun Gong, Nan Duan, and Ruofei Zhang.
\newblock El-attention: Memory efficient lossless attention for generation.
\newblock In {\em International Conference on Machine Learning}, pages 11648--11658. PMLR, 2021.

\bibitem{yang2025qwen3technicalreport}
An~Yang, Anfeng Li, Baosong Yang, Beichen Zhang, Binyuan Hui, Bo~Zheng, Bowen Yu, Chang Gao, Chengen Huang, Chenxu Lv, Chujie Zheng, Dayiheng Liu, Fan Zhou, Fei Huang, Feng Hu, Hao Ge, Haoran Wei, Huan Lin, Jialong Tang, Jian Yang, Jianhong Tu, Jianwei Zhang, Jianxin Yang, Jiaxi Yang, Jing Zhou, Jingren Zhou, Junyang Lin, Kai Dang, Keqin Bao, Kexin Yang, Le~Yu, Lianghao Deng, Mei Li, Mingfeng Xue, Mingze Li, Pei Zhang, Peng Wang, Qin Zhu, Rui Men, Ruize Gao, Shixuan Liu, Shuang Luo, Tianhao Li, Tianyi Tang, Wenbiao Yin, Xingzhang Ren, Xinyu Wang, Xinyu Zhang, Xuancheng Ren, Yang Fan, Yang Su, Yichang Zhang, Yinger Zhang, Yu~Wan, Yuqiong Liu, Zekun Wang, Zeyu Cui, Zhenru Zhang, Zhipeng Zhou, and Zihan Qiu.
\newblock Qwen3 technical report, 2025.

\bibitem{yang2024no}
June~Yong Yang, Byeongwook Kim, Jeongin Bae, Beomseok Kwon, Gunho Park, Eunho Yang, Se~Jung Kwon, and Dongsoo Lee.
\newblock No token left behind: Reliable kv cache compression via importance-aware mixed precision quantization.
\newblock {\em arXiv preprint arXiv:2402.18096}, 2024.

\bibitem{yuan2024llm}
Zhihang Yuan, Yuzhang Shang, Yang Zhou, Zhen Dong, Zhe Zhou, Chenhao Xue, Bingzhe Wu, Zhikai Li, Qingyi Gu, Yong~Jae Lee, et~al.
\newblock Llm inference unveiled: Survey and roofline model insights.
\newblock {\em arXiv preprint arXiv:2402.16363}, 2024.

\bibitem{zhang2019rootmeansquarelayer}
Biao Zhang and Rico Sennrich.
\newblock Root mean square layer normalization.
\newblock In {\em Proceedings of the 33rd International Conference on Neural Information Processing Systems}, pages 12381--12392, 2019.

\bibitem{zhang2024lorc}
Rongzhi Zhang, Kuang Wang, Liyuan Liu, Shuohang Wang, Hao Cheng, Chao Zhang, and Yelong Shen.
\newblock Lorc: Low-rank compression for llms kv cache with a progressive compression strategy.
\newblock {\em arXiv preprint arXiv:2410.03111}, 2024.

\end{thebibliography}

\appendix
\counterwithin{figure}{section}
\counterwithin{table}{section}
\appendix
\clearpage
\onecolumn

\section{Prefill for \OURSCL}
\label{sec:appendix-xquant-cl}

Here we include a visualization of the prefill phase of inference for \OURSCL. In particular, we show how the deltas between the $X$ embeddings for successive layers are calculated, quantized, and cached. Note that besides Layer 0's input $X_0$, the input for all other layers is an approximation calculated using the sum of $X_0$ and the previous quantized deltas.

\begin{figure}[h]
\centering
\includegraphics[width=1\linewidth]{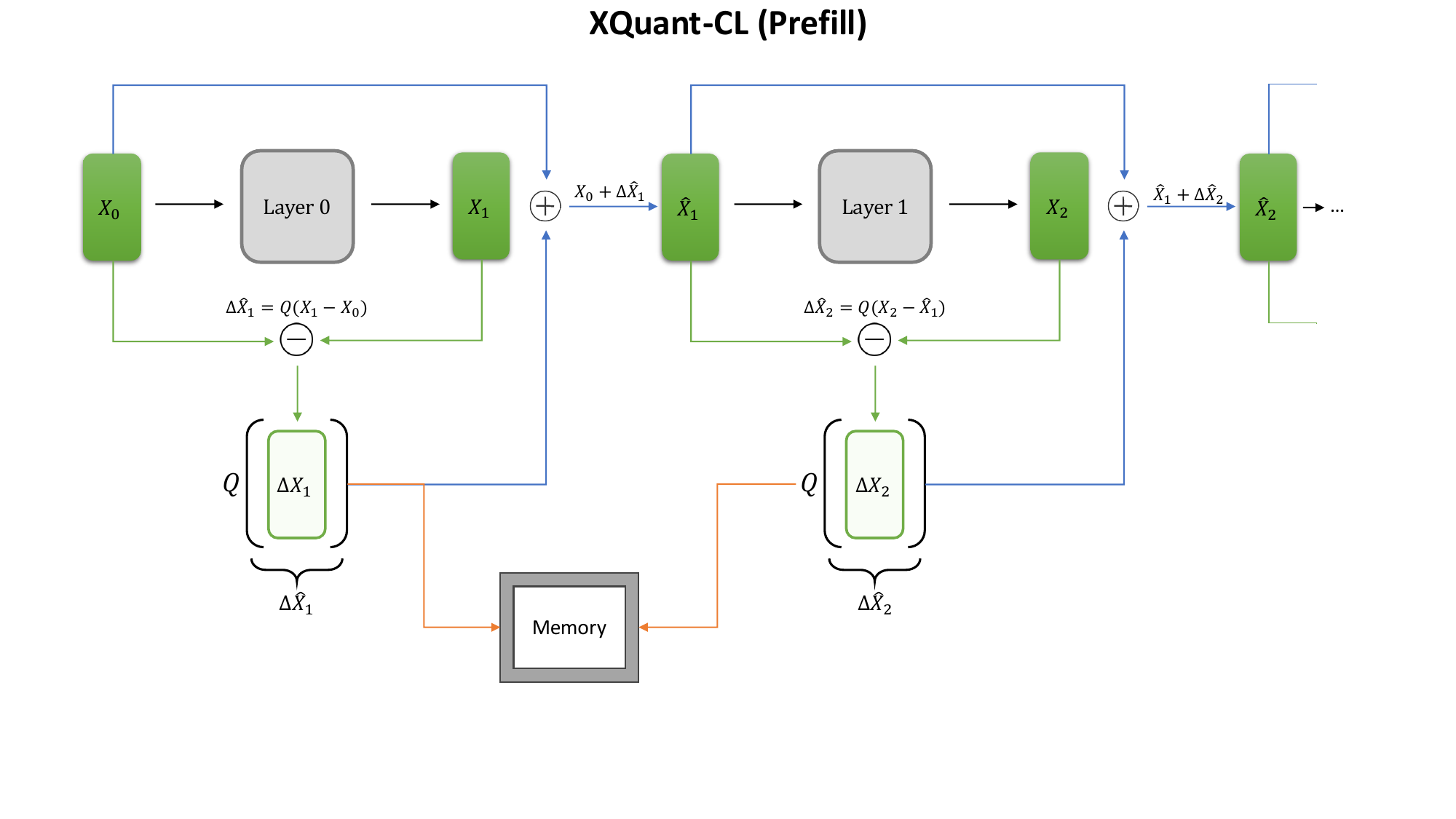}
 \caption{
 Illustration of \OURSCL algorithm during prefill. Besides Layer 0, the input to all other layers is a cross layer approximation, computed using the deltas of all previous layers and the input of Layer 0. The input of Layer 0 is summed with each layer's delta so it can be treated as an accumulator, allowing us to avoid loading all $N-1$ deltas to compute Layer $N$'s $X$. After a layer is done processing, the input embeddings to the layer for all tokens in the sequence are subtracted from the output activations of the layer (the same shape as the input embeddings), and this delta is quantized and cached as $\Delta \hat{X}$.
}
  \label{fig:cross-layer-prefill}
\end{figure}

\section{Observed Outlier Property When Applying \OURS for GQA Models}
\label{sec:appendix-xquant-gqa}

\begin{figure}[t]
    \includegraphics[width=1\linewidth]{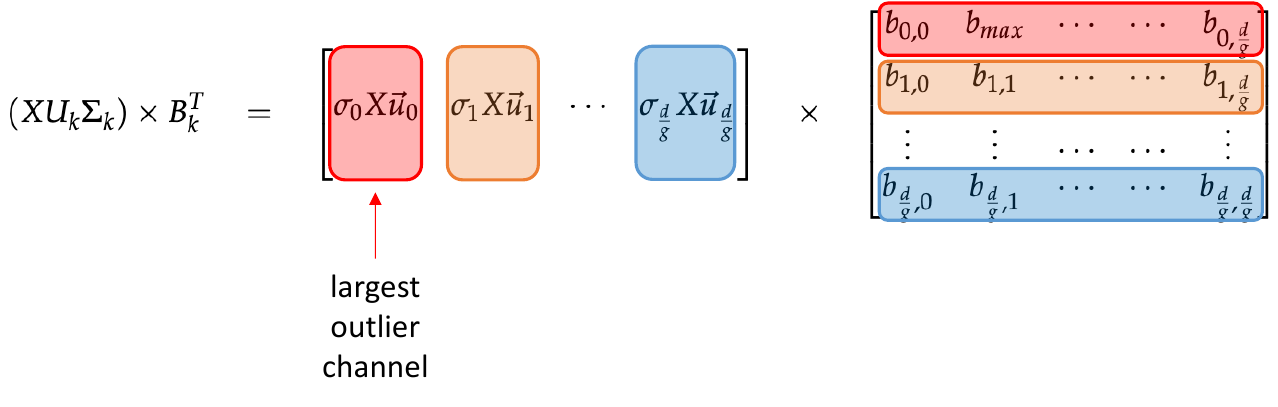}
    \vspace{1em}
    \caption{
    $X \in \mathbb{R}^{L\times d}$, where $L$ is the sequence length, and $d$ is the hidden dimension. $U_k \in \mathbb{R}^{d \times \frac{d}{g}}$ and $\Sigma_k \in \mathbb{R}^{\frac{d}{g}, \frac{d}{g}}$} where $g$ is the group size used by GQA models. Note that $\sigma_0 \ge \sigma_1 \ge \cdots \geq \sigma_{\frac{d}{g}} > 0$. We find that $XU_k$ has massive outliers in the first channel across all layers for different models on different datasets (see Figures \ref{fig:llama-latent-x-distributions}, \ref{fig:mistral-latent-x-distributions}). Multiplying $XU_k$ by $\Sigma_k$ preserves the ordering of outlier channels (first channel still has the largest magnitude outliers). When multiplying $XU_k\Sigma_k$ by $B^T_v$, the first row of $B^T_v$ interacts with the first outlier channel of $XU_k\Sigma_k$, as highlighted in red above. The top-k largest magnitude values from this first row can be used to identify which channels in the Keys are outlier channels. This can be achieved offline without any calibration data, and simply by inspecting the weights of $B^T_v$.
    \label{fig:xquant_kvquant_outlier_connection}
\end{figure}

\paragraph{Recap} As a reminder, we decompose the Key and Value projection matrices $W_k$ and $W_v$ using the SVD to get $U_k\Sigma_kB^T_k$ and $U_v\Sigma_vB^T_v$ respectively. We down-project $X$ by $U_k$ and $U_v$ and quantize and cache these latent distributions, while also fusing $\Sigma_kB^T_k$ into one weight matrix and $\Sigma_vB^T_v$ into another weight matrix, both of which respectively multiply $U_k$ and $U_v$ to rematerialize the Keys and Values.

\begin{figure}[t]
    \vspace{-5em}
    \centering
    \includegraphics[width=0.8\linewidth]{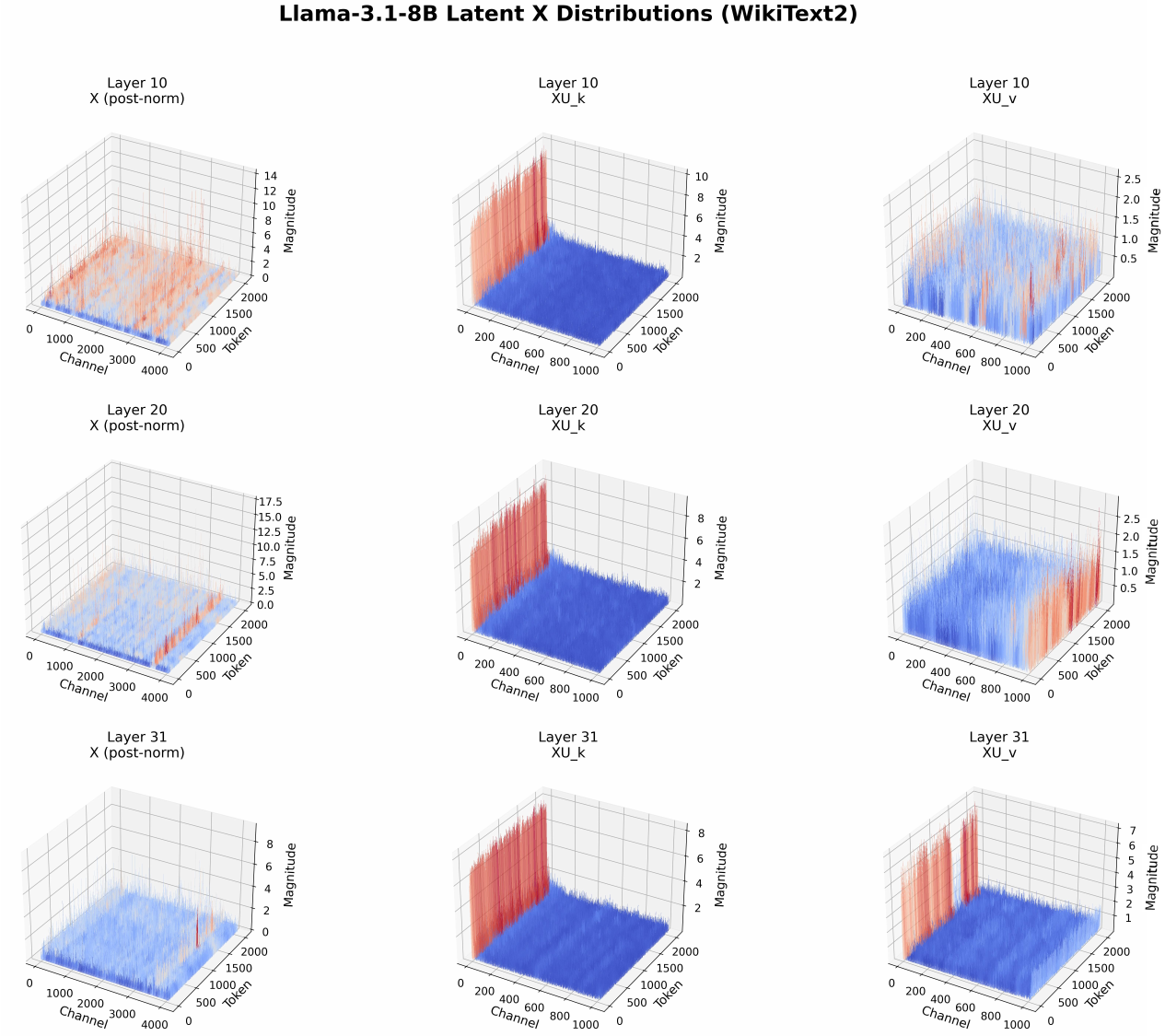}
    \vspace{2em}
    
    \includegraphics[width=0.8\linewidth]{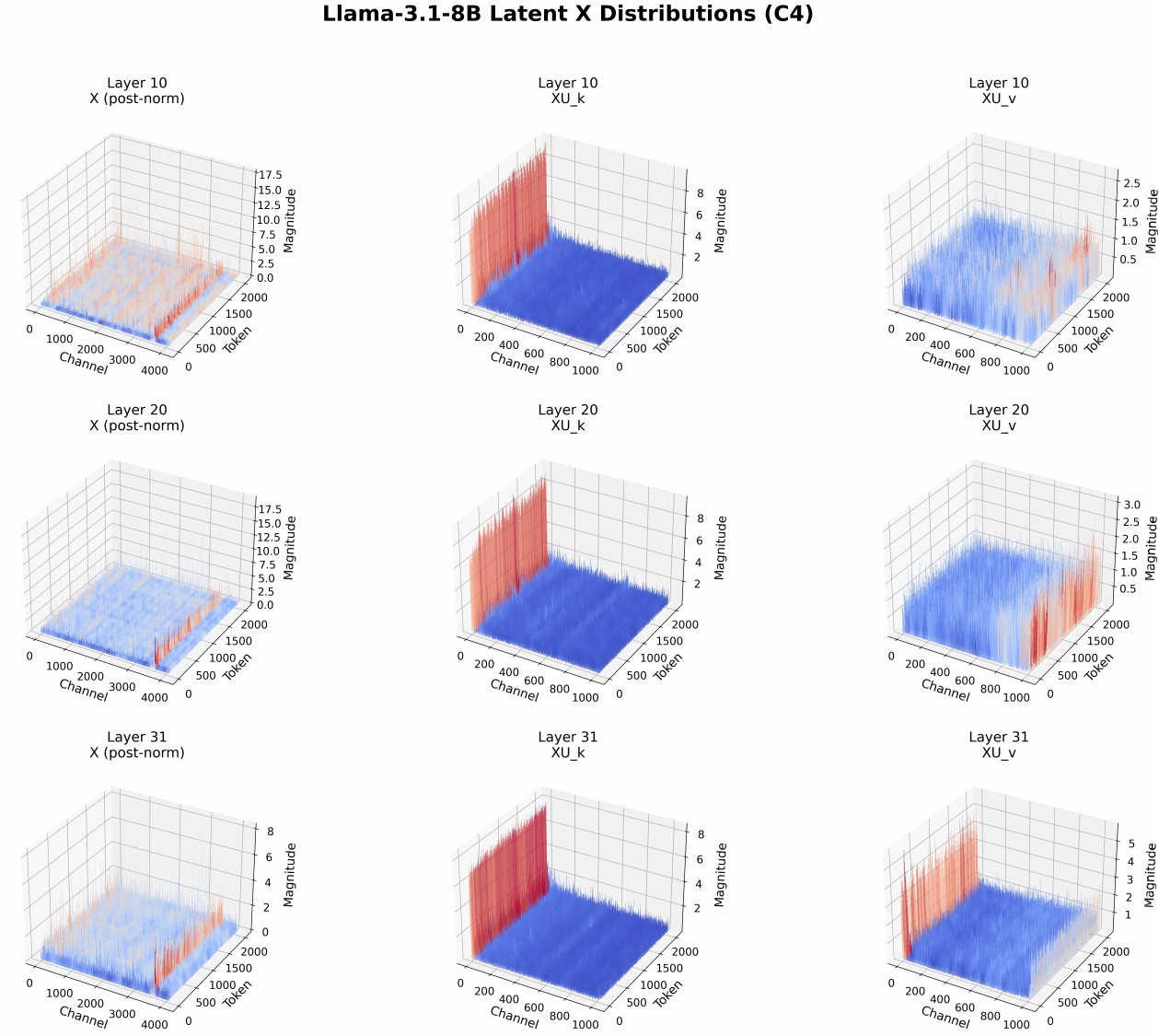}
    \caption{$X$ (left), $XU_k$ (middle), and $XU_v$ (right) distributions for Llama-3.1-8B on samples from WikiText-2 (top) and C4 (bottom).}
    \label{fig:llama-latent-x-distributions}
\end{figure}

\begin{figure}[t]
    \vspace{-5em}
    \centering
    \includegraphics[width=0.8\linewidth]{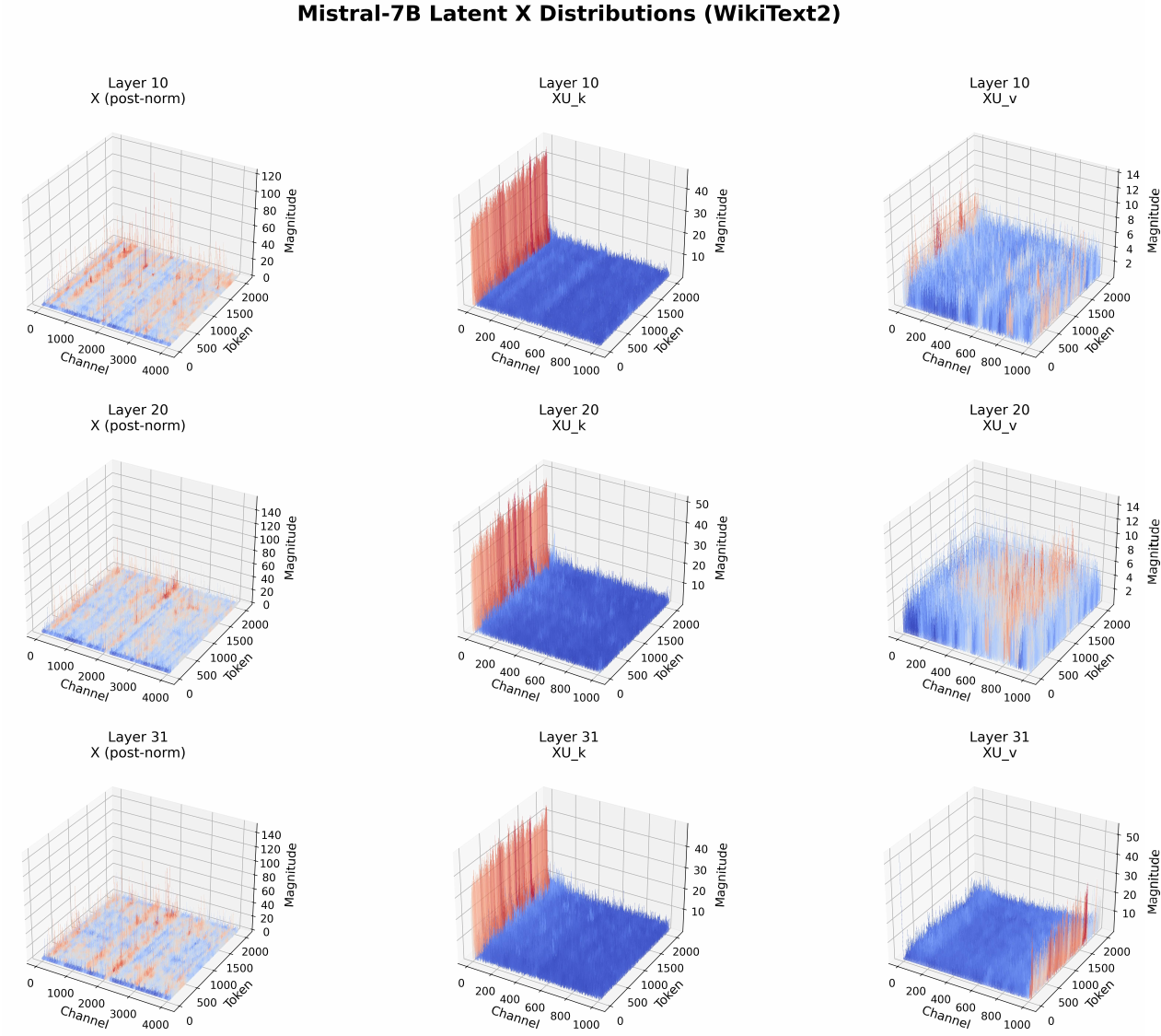}
    \vspace{2em}

    \includegraphics[width=0.8\linewidth]{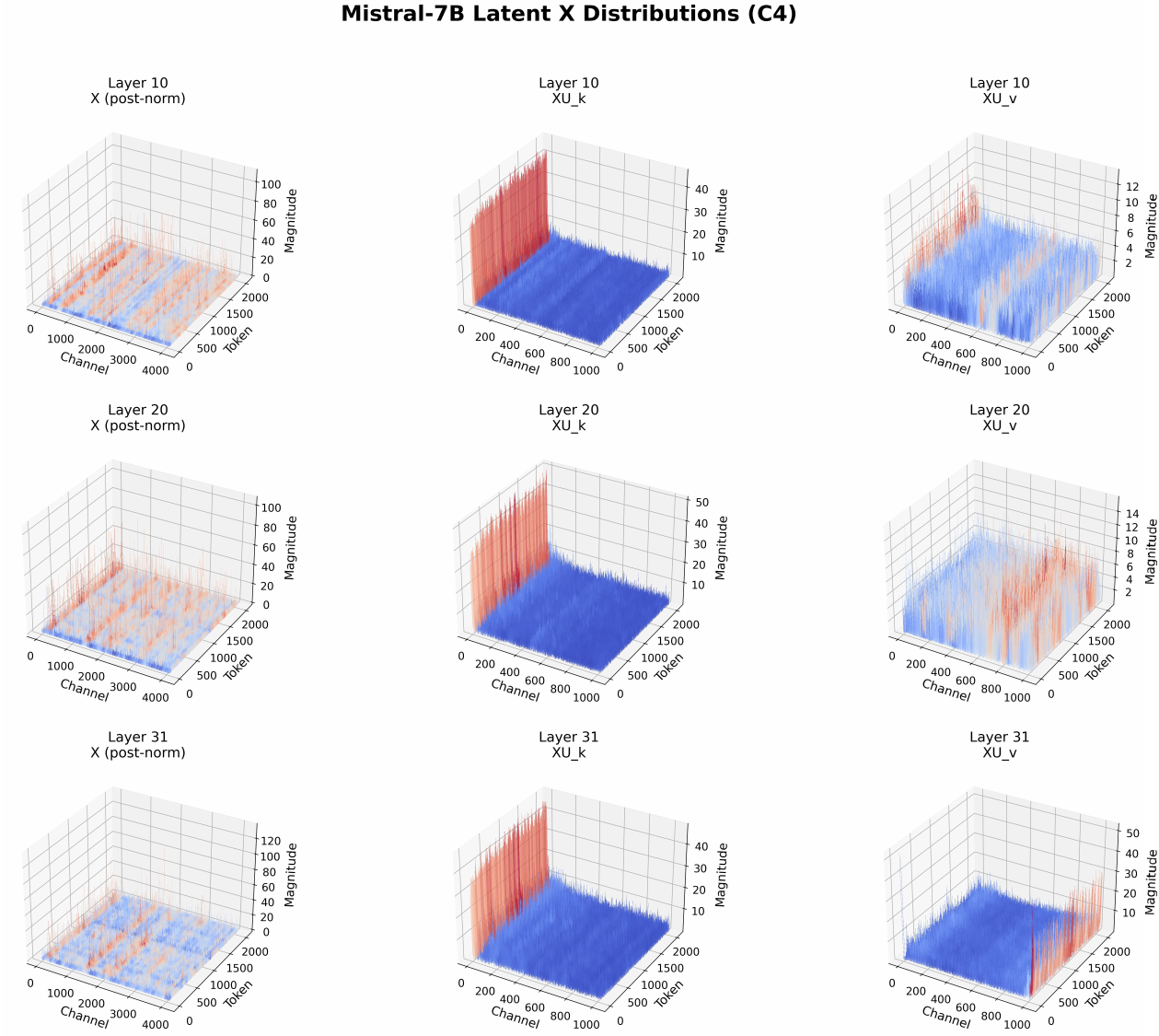}
    \caption{$X$ (left), $XU_k$ (middle), and $XU_v$ (right) distributions for Mistral-7B on samples from WikiText-2 (top) and C4 (bottom).}
    \label{fig:mistral-latent-x-distributions}
\end{figure}

\paragraph{Observed Outlier Property} When we project $X$ by $U_k$, we observe an interesting property of these distributions across all layers for all models, where $U_k$ transforms $X$ such that all outliers are grouped on the first channel of the resulting matrix. Although we observe this phenomenon for both MHA and GQA models, we discuss it here in the context of GQA models. This is because down-projecting $X$ by $U_k$ gives us memory savings and is relevant to the $\OURS$ algorithm for GQA models, whereas for MHA models, we simply cache $X$ itself. We visualize the $X$, $XU_k$, and $XU_v$ distributions for the Llama-3.1-8B and Mistral-7B models in Figures \ref{fig:llama-latent-x-distributions} and \ref{fig:mistral-latent-x-distributions} respectively. For each model, we visualize the distributions for 3 different layers on the WikiText-2 and C4 datasets to ensure that this observed property cannot simply be attributed to a specific dataset, model, or layer. Note that these distributions \emph{do not include} the scaling term $\Sigma$ from the SVD -- they only visualize each of the transformations that the left singular vectors matrices $U_k$ and $U_v$ with orthonormal columns apply on the input $X$.

As mentioned in \ref{sec:xquant-gqa}, we notice a 
pattern in the $XU_k$ distributions across all layers of both models, where the first channel has massive outliers across all tokens, whereas all other channels are much smaller in magnitude. We do not, however, observe any similar interesting quality for the $XU_v$ distributions. This observation implies that the activation embeddings for all tokens have high cosine similarity with the first column vector of $U_k$, which is also the left singular vector associated with the largest singular value of $W_k$. In other words, the activation embeddings for all tokens roughly lie close to the direction where $W_k$ applies maximum scaling.

We attempt to exploit this finding by keeping the first outlier channel of $XU_k$ in FP16, while quantizing the remaining channels using per-channel quantization with group-size 128 as before. We use the same setup as discussed in Section \ref{sec:results}. These results are shown in Table \ref{tab:appendix-outlier-channel-fp16}. The data for KIVI* and \OURS have been duplicated from Table \ref{tab:xquant-ppl} to serve as a reference point. We find that keeping the outlier channel in FP16 gives non-trivial benefits for 2-bit quantization. For instance, Llama-3.1-8B sees an improvement of 0.2 perplexity on C4 in 2-bit precision.

\paragraph{Connection to KV cache quantization methods} Previous KV cache quantization methods such as KVQuant \cite{hooper2024kvquant} preserve outlier values in the Keys in higher precision to reduce accuracy degradation from quantization. Since the Keys have been observed to have distinct outlier channels (see Figure \ref{fig:cross-layer-distn}), determining which outliers to preserve in the entire Keys matrix requires finding these outlier channels \cite{liu2024kivi, hooper2024kvquant}. To identify these outlier channels, methods like \cite{hooper2024kvquant} run models on calibration datasets, which allows them to note which channels tend to be outliers across diverse data samples.

Having observed the outlier behavior of the first channel in $XU_k$, we explore if outlier channels in the Keys can be identified by inspecting the SVD decomposition of the $W_k$ matrix offline without the need to run calibration. We follow this line of reasoning: $XU_k$ is an ordered distribution where the first channel (i.e., first column) is associated with the largest singular value of $W_k$, the second channel is associated with the second largest singular value, and so on. If we multiply $XU_k$ by $\Sigma_k$, the first channel in $XU_k$ gets scaled by the largest singular value $\sigma_0$, the second channel in $XU_k$ gets scaled by the second largest singular value $\sigma_1$, and so on. This matrix multiplication is visualized in Figure \ref{fig:xquant_kvquant_outlier_connection}. Thus, $XU_k\cdot\Sigma_k$ preserves the ordering of the outlier channels (i.e., the first channel in $XU_k$ is still the outlier channel). Finally to complete the rematerialization of Keys, $(XU_k\Sigma_k)$ is multiplied by $B^T_v$, which necessarily distributes the outliers in the first channel to other channels. These are the channels that KV cache quantization methods try to identify in the Keys distribution via running calibration. 

However, we hypothesize that the column in $B^T_v$ whose first element has the largest magnitude compared to the first element of all other columns will also tend to be the column of the Keys containing the outlier channel. For example, in Figure \ref{fig:xquant_kvquant_outlier_connection}, the second column of $B_v^T$ is shown to contain $b_{max}$, which is the largest magnitude value in the first row of $B^T_v$. This means that the second column of the Keys will tend to be the outlier channel. This is because the values in the first row vector of $B^T_v$ are the scalars which interact with the first outlier channel in $(XU_k\Sigma_k)$ when performing matrix multiplication. So, the column in the first row vector of $B^T_v$ with the largest scalar that "hits" the first column in $(XU_k \Sigma_k)$ will result in the largest values in the Keys. In Figure \ref{fig:xquant_kvquant_outlier_connection}, the largest outlier channel $\sigma_0 X \vec{u}_0$ is highlighted in red, and the first row vector in $B^T_v$ containing the scalars which multiply this outlier channel are also highlighted in red.

However, there are two cases in which the above hypothesis does not hold: 1) other row vectors in $B^T_v$ contain scalars which cause another channel that is not the first channel in $XU_k$ to blow up in magnitude and 2) the largest scalar of $B^T_v$'s first row vector results in most values of the outlier channel having an opposite sign to the next largest outlier channel of $(XU_k\Sigma_k)$, causing their sum to have smaller magnitude. Nevertheless, these scenarios are less likely given that the other channels are themselves much smaller in magnitude than the first channel, so it is unlikely that 1) any other channel could blow up to surpass the first outlier channel in magnitude and 2) adding the first outlier channel to another channel whose elements are the opposite sign will impact the first outlier channel much.

We test this hypothesis by using the above method to determine the outlier channels in the Keys, and then check how many of these predictions are correct by comparing with the ground truth. For the ground truth, we pick the channel in the Keys matrix which has the largest average magnitude. As there can be multiple outlier channels in the Keys, we choose the column indices for the top-k largest magnitude values in the first row of $B^T_v$ as our predictions. We then simply check if the ground truth index for the outlier channel in the Keys appears in any of the predicted indices by the above method. We mark the predictions as correct if any one of them contain the ground truth index, and we report the final accuracy as the percentage of correct predictions across the Keys for all layers of the model. We test this method on two different datasets, WikiText-2 and C4, to ascertain whether this weights-only based analysis is robust to different data. We also test on the Llama-3.1-8B and Mistral-7B models to check the method's efficacy across different models. Results are listed in Table \ref{tab:appendix-xquant-outlier-approx}.

We find that for Llama-3.1-8B, only 8 of the largest magnitude values of the first of $B^T_v$ are needed to determine the outlier channel for the Keys. For Mistral-7B, using the top-8 largest magnitude values attains 96.88\% accuracy. Moreover, for both models, the accuracy is consistent across datasets, demonstrating the robustness of this method to determine the outlier indices for the Keys.

\begin{table}[t]
\centering
\caption{We attempt to exploit the structured distribution of $XU_k$ by keeping the first outlier channel in FP16. The table shows \OURS evaluation using perplexity on WikiText-2 and C4 on Llama‑3.1‑8B and Mistral‑7B. We duplicate the KIVI* and \OURS results from Table \ref{tab:xquant-ppl} to serve as a reference point. For 2-bit precision, keeping the first channel in FP16 results in some perplexity improvement.}
\label{tab:appendix-outlier-channel-fp16}
\footnotesize
\renewcommand{\arraystretch}{1.2}
\setlength{\tabcolsep}{6pt}
\begin{tabular}{c|c|cc|cc}
\toprule
\multirow{2}{*}{\textbf{Method}} &
\textbf{KV} &
\multicolumn{2}{c|}{\textbf{Llama‑3.1‑8B}} &
\multicolumn{2}{c}{\textbf{Mistral‑7B}} \\
\cmidrule{3-6}
& \textbf{(GQA)} & Wiki2 & C4 & Wiki2 & C4 \\
\midrule\midrule
Baseline            & 1.00 & 6.24 & 9.54  & 5.32 & 8.47 \\
\midrule
KIVI*-4bit          & 0.27 & 6.31 & 9.66  & 5.34 & 8.51 \\
\OURS-4bit      & 0.27 & 6.28 & 9.60  & 5.33 & 8.49 \\
\hc \OURS-4bit-FP16-outlier-channel     & 0.27 & \textbf{6.28} & \textbf{9.60}  & \textbf{5.33} & \textbf{8.49} \\
\midrule
KIVI*-3bit          & 0.20 & 6.59 & 10.13 & 5.43 & 8.62 \\
\OURS-3bit      & 0.20 & 6.43 & 9.89  & 5.39 & 8.57 \\
\hc \OURS-3bit-FP16-outlier-channel     & 0.20 & \textbf{6.42} & \textbf{9.87}  & \textbf{5.38} & \textbf{8.56} \\
\midrule
KIVI*-2bit          & 0.14 & 9.95 & 15.98 & 6.36 & 9.88 \\
\OURS-2bit      & 0.14 & 7.74 & 12.27 & 5.79 & 9.13 \\
\hc \OURS-2bit-FP16-outlier-channel     & 0.14 & \textbf{7.63} & \textbf{12.05}  & \textbf{5.75} & \textbf{9.08} \\
\bottomrule
\end{tabular}
\end{table}

\begin{table}[h]
\caption{
Percentage of outlier channels predicted correctly by only analyzing the top-k values of the weight matrix $B^T_v$ offline, without any calibration data. For Llama-3.1-8B, only 8 of the largest magnitude values of the first row of $B^T_v$ are needed to determine the outlier channel for the Keys.
}
\label{tab:appendix-xquant-outlier-approx}
\vspace{1mm}
\small
\centering
\renewcommand{\arraystretch}{1.2}
\setlength{\tabcolsep}{6pt}
\begin{tabular}{c|cc|cc}
\toprule
\textbf{top-k} &
\multicolumn{2}{c|}{\textbf{Llama‑3.1‑8B}} &
\multicolumn{2}{c}{\textbf{Mistral‑7B}} \\
\cmidrule{2-5}
& WikiText2 & C4 & WikiText2 & C4 \\ \midrule\midrule
k=1     & 66.14\% & 71.15\% & 75.35\% & 71.91\% \\
\midrule
k=2     & 72.08\% & 75.08\% & 87.55\% & 83.48\% \\
\midrule
k=4     & 87.71\% & 90.62\% & 93.75\% & 93.75\% \\
\midrule
\hc k=8     & \textbf{100\%} & \textbf{100\%} & \textbf{96.88}\% & \textbf{96.88\%} \\
\bottomrule
\end{tabular}
\end{table}

\end{document}